\documentclass[11pt]{article}

\usepackage[final]{acl}

\usepackage{times}
\usepackage{latexsym}
\usepackage[T1]{fontenc}
\usepackage[utf8]{inputenc}
\usepackage{microtype}
\usepackage{inconsolata}

\usepackage[utf8]{inputenc} 
\usepackage[T1]{fontenc}    
\usepackage{multirow}
\usepackage{amsmath}
\usepackage{graphicx}
\usepackage{hyperref} 
\usepackage{algorithm}
\usepackage{algorithmic}
\usepackage{microtype}
\usepackage{url}
\usepackage{booktabs}
\usepackage{amsfonts}
\usepackage{bm}
\usepackage{enumitem}
\usepackage[most]{tcolorbox}
\usepackage{xcolor}
\usepackage{adjustbox}
\usepackage{wrapfig}
\usepackage[most]{tcolorbox}
\usepackage[nameinlink]{cleveref}
\crefname{appendix}{Appendix}{Appendices}
\Crefname{appendix}{Appendix}{Appendices}
\usepackage{booktabs}
\usepackage[dvipsnames]{xcolor}
\usepackage{amssymb}
\usepackage{parskip}
\usepackage{titletoc}

\definecolor{darkblue}{rgb}{0, 0, 0.5}
\hypersetup{colorlinks=true, citecolor=blue, linkcolor=darkblue, urlcolor=darkblue}

\usepackage{color-edits}
\addauthor{kw}{orange}

\newcommand{\mname}{\texttt{ClinicalReTrial}}

\title{\mname: Clinical Trial Redesign with Self-Evolving Agents}

\author{
  \textbf{Sixue Xing\textsuperscript{1}},
  \textbf{Kerui Wu\textsuperscript{2}},
  \textbf{Xuanye Xia\textsuperscript{3}},
  \textbf{Haoyu He\textsuperscript{4}},
  \textbf{Meng Jiang\textsuperscript{1}},
  \textbf{Jintai Chen\textsuperscript{5}},
  \textbf{Tianfan Fu\textsuperscript{6}}
  \\
  \\
  \textsuperscript{1}University of Notre Dame, Notre Dame, IN, USA \\
  \textsuperscript{2}University of Massachusetts Amherst, Amherst, MA, USA \\
  \textsuperscript{3}Georgia Institute of Technology, Atlanta, GA, USA \\
  \textsuperscript{4}Northeastern University, Boston, MA, USA \\
  \textsuperscript{5}Hong Kong University of Science and Technology (Guangzhou), Guangzhou, Guangdong, China \\
  \textsuperscript{6}State Key Laboratory for Novel Software Technology at Nanjing University, Nanjing, Jiangsu, China
}

\begin{document}

\maketitle
    
\begin{abstract}
Clinical trials constitute a critical yet exceptionally challenging and costly stage of drug development (\$2.6B per drug), where protocols are encoded as complex natural language documents, motivating the use of AI systems beyond manual analysis. Existing AI methods accurately predict trial failure, but do not provide actionable remedies. To fill this gap, this paper proposes \mname, a multi-agent system that formulates clinical trial optimization as an iterative redesign problem on textual protocols. Our method integrates failure diagnosis, safety-aware modifications, and candidate evaluation in a closed-loop, reward-driven optimization framework. 
Serving the outcome prediction model as a simulation environment, \mname\ enables low-cost evaluation and dense reward signals for continuous self-improvement. We further propose a hierarchical memory that captures iteration-level feedback within trials and distills transferable redesign patterns across trials. Empirically, \mname\ turns 56.7\% of failed protocols into predicted successes under the simulation environment, with a mean success probability gain of 7.4\% at negligible cost (\$0.156 per trial). Extensive retrospective case studies further show that \mname\ recovers clinically meaningful modification patterns that align with real-world expert-driven protocol changes.
The code is available at: \url{https://github.com/xingsixue123/ClinicalRetrial}.
\end{abstract}

\vspace{-7pt}
\section{Introduction}\label{sec:intro}
Clinical trials represent the most critical and expensive phase in drug discovery, with an estimated cost of \$2.6 billion~\citep{dimasi2016innovation} per approved drug, and low success rates of approximately 10-20\%~\citep{yamaguchi2021approval}. Serving as documented plan that specifies the study's objectives, clinical trial protocols involve complex, interdependent design choices \citep{getz2017trends} expressed in natural language documents, such as eligibility criteria, dosing strategies, and endpoint definitions, where small design flaws can propagate into irreversible failure. These challenges motivate the use of AI systems \cite{zhang2023harnessing} that can reason over high-dimensional trial designs, leverage historical evidence, and systematically assess failure risks at scale.

Recent advances in AI have enabled increasingly accurate prediction of clinical trial outcomes. For example, \citet{lo2019machine} uses structured metadata to model success likelihood; \citet{fu2022hint, chen2024uncertainty, chen2025trialbench} integrate heterogeneous data sources using architectures including graph neural networks and hierarchical attention mechanisms to predict trial approval rate; \citet{yue2024clinicalagent, liu2025autoct} incorporate Large Language Models (LLMs) and external knowledge bases to enhance reasoning and explainability in trial outcome prediction through advanced natural language understandings of complex medical texts.

Despite their success, existing approaches are inherently reactive in nature: they operate on a fixed clinical trial protocol and produce a prediction or post-hoc explanation of trial success or failure. However, these methods do not address a more practically consequential problem: they are unable to respond to a trial failure due to the lack of actionable interventions. In real-world drug discovery, stakeholders require not only assessments of failure risk, but also actionable guidance on protocol redesign, including principal modifications or enhancements informed by identified sources of risk~\citep{dagenais2022use,baumfeld2020trial}.

\begin{figure*}[ht]
    \centering
    \includegraphics[width=1\linewidth]{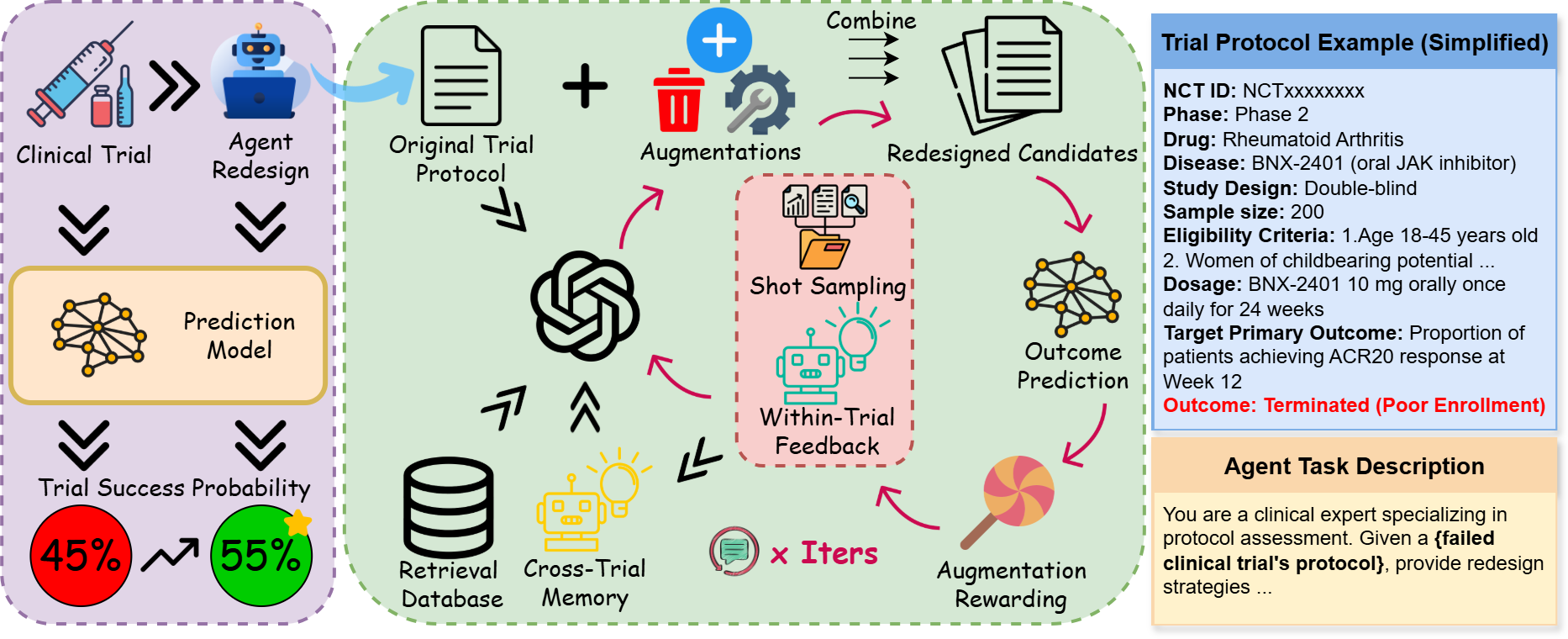}
    \caption{\mname\ Agent architecture. The system operates through iterative refinement: agents analyze failures, generate modifications, and receive rewards from the simulation environment. Historical explorations are extracted into structured knowledge that guides subsequent iterations, enabling progressive improvement.}
    \label{fig:retrial_agent}
    \vspace{-10pt}
\end{figure*}

To bridge this gap, in this work, as shown in \Cref{fig:retrial_agent}, we propose \mname, a self-evolving AI agent that moves beyond static prediction toward actionable intervention through the optimization of the trial protocol, while continuously improving its redesign policies. The language-rich nature of clinical trial protocols makes this an ideal domain for LLM-based optimization. Our framework instantiates a coordinated multi-agent pipeline that performs failure diagnosis, protocol redesign, and candidate evaluation, with domain knowledge and safety awareness embedded at each decision stage. Beyond a single optimization run, we adopt the prediction model as a simulation environment to provide rewards for continuous self-improvement. Specifically, \mname\ maintains local memory to accumulate iteration-level feedback and reward attributed modification outcomes for within-trial adaptation, while a global memory distills transferable redesign patterns across trials to enable warm start initialization and exploration calibration. Through this hierarchical learning structure and reward-driven closed-loop optimization, \mname\ systematically explores the protocol modification space and learns to identify high-impact interventions that improve clinical trial success probability.

Experimentally, our prediction model demonstrated the strongest performance (PR-AUC$>0.75$), allowing it to serve as a reliable simulation environment for evaluation and agent optimization. In the trial redesign experiments, \mname\ successfully improved trial protocols with a mean success probability gain $\Delta p = 7.4\%$, which turns 56.7\% of failed protocols into predicted successes under the simulation environment, at negligible cost ($\$0.156/$trial). 
Moreover, across 8 real-world retrospective case studies, \mname\ rediscovers clinically meaningful redesign strategies from failed protocols, producing modifications that closely align with independently successful expert-driven trial amendments.

\paragraph{Main contributions}
(1) (to the best of our knowledge) We are the first to formulate clinical trial optimization as an AI-solvable and \textit{in silico} verifiable problem.
(2) We propose a multi-agent pipeline with domain knowledge that decomposes clinical trial protocol optimization into diagnosis, modification, and evaluation.
(3) We develop a simulation-driven clinical trial optimization framework 
with \textit{In-Context Learning} and multi-level memory for 
self-improvement through reward-attributed prompt optimization, 
dynamic redesign pool curation, and cross-trial knowledge distillation.

\vspace{-2pt}
\section{Related Work}\label{sec:related_work}

Early efforts employed classical machine learning (logistic regression (LR), random forests) on expert-curated features~\citep{gayvert2016clinical,lo2019machine}, establishing feasibility but lacking multi-modal data integration. Deep learning approaches addressed this: \citet{fu2022hint,fu2023automated} proposed HINT, integrating drug molecules, ICD-10 codes, and eligibility criteria; \citet{chen2024uncertainty,lu2024uncertainty} added uncertainty quantification and interpretability; \citet{wang2024twin,yue2024trialenroll} designed LLM-based patient-level digital twins; \citet{chen2025trialbench} released a standardized TrialBench with multi-modal baselines while maintaining competitive performance. Recent LLM approaches demonstrate medical reasoning~\citep{singhal2023medpalm}, enhanced via retrieval-augmented generation~\citep{lewis2020rag} with databases like DrugBank~\citep{wishart2018drugbank}, Hetionet~\citep{himmelstein2017hetionet}, and domain-adapted encoders like BioBERT~\citep{lee2020biobert}. Building on this, ClinicalAgent~\citep{yue2024clinicalagent} decomposes prediction into specialized sub-task agents with ReAct reasoning~\citep{yao2023react}. \citet{liu2025autoct} proposed AutoCT for autonomous feature engineering via Monte Carlo Tree Search~\citep{chi2024sela}. 
However, these methods are essentially predictive models without explaining \textit{why} failures occur or \textit{how} to modify protocols. 
In contrast, our multi-agent architecture leverages chain-of-thought~\citep{wei2022chain} and least-to-most prompting~\citep{zhou2023least} to dynamically diagnose failure causes and refine protocols.
\section{Methodology}\label{sec:method}
\noindent\textbf{Overview}. \mname\ is a self-improving multi-agent system that redesigns failed clinical trials through reward-driven iterative optimization. We first formulate the clinical trial optimization problem in \Cref{sec:problem_identification}. Then, we build a multi-agent framework to address it in \Cref{agent_pipeline}. To further improve our Agent performance, we integrate a knowledge retrieval system and present In-Context Learning with multi-level memory in \Cref{sec:hierarchical_learning}. For ease of exposition, \Cref{fig:agent_workflow} illustrates the process, and \Cref{alg:main} provided in Appendix~\ref{app:algo} formalizes the iterative optimization procedure.

\begin{figure*}[ht]
    \centering
    \includegraphics[width=1\linewidth]{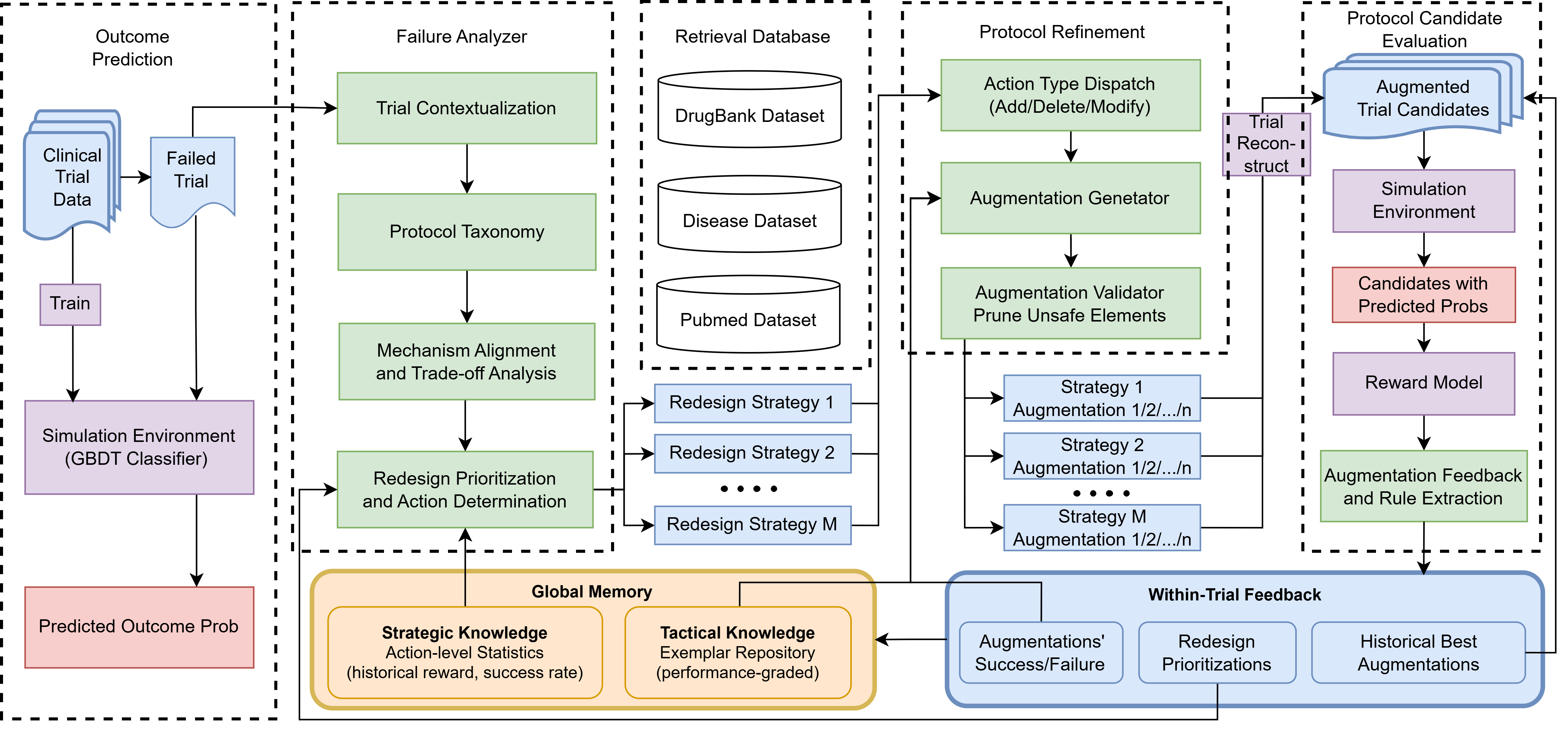}
    \caption{Detailed \mname\ Agent architecture and iterative redesign workflow, comprising diagnosis, augmentation, and validation, with a simulation environment and hierarchical memory to progressively optimize failed clinical trial protocols.}
    \label{fig:agent_workflow}
\end{figure*}

\subsection{Problem: Clinical Trial Optimization}
\label{sec:problem_identification}
The goal of clinical trial is to evaluate the safety and efficacy of drug on patients. 
A clinical trial protocol, including eligibility criteria and drug dosage specifications, defines the study's framework in a standardized and controlled manner.
Formally, let $T_0 = \{e_1, e_2, \ldots, e_K\}$ denote a clinical trial protocol decomposed into $K$ modifiable elements (\textit{e.g.}, eligibility criteria, dosage regimens, endpoint definitions), where a prediction model $f_{\theta}: \mathcal{T} \rightarrow [0, 1]$ assigns success probability $p_0 = f_{\theta}(T_0)$. Each element $e_i$, after redesign, admits a set of augmentations $\mathcal{A}_i = \{a_{i1}, a_{i2}, \ldots, a_{im_i}\}$ representing clinically valid modifications, where candidate protocols $\mathbb{T} = \{T'_1, T'_2, \ldots, T'_N\}$ are reconstructed by selectively replacing elements in $T_0$ with their augmented variants. Such an exploration set of redesigns $\mathbb{T}$ is evaluated through the prediction model to obtain success probabilities, with each $p'_j = f_{\theta}(T'_j)$. The optimization objective seeks the optimal protocol $T^* = \arg\max_{T' \in \mathbb{T}} f_{\theta}(T')$ with success probability $p^*$, where overall improvement is measured by $\Delta p = p^* - p_0$ with $p^* = f_{\theta}(T^*)$.

\subsection{Multi-Agent Architecture}
\label{agent_pipeline}


With heterogeneous necessities, clinical trial redesign requires distinct diagnostic expertise and remediation strategies \cite{fogel2018factors}. We introduce a multi-agent architecture \cite{chen2024survey} that naturally aligns with this structure as a closed loop performing typed editing over the protocol's textual constraints: the Trial Failure Analyzer (§\ref{sec:analysis_agent}) diagnoses root causes and proposes typed edit operations; the Protocol Refinement Generator (§\ref{sec:augment_agent}) realizes them as concrete protocol text; the Clinical Safety Validator (§\ref{sec:validation_agent}) filters them against external domain knowledge; and the Protocol Candidate Evaluator (§\ref{sec:exploration_orchestrator}) scores complete candidates with simulation-based feedback.

\subsubsection{Protocol Diagnosis Analyzer}
\label{sec:analysis_agent}

Given a failed protocol and prior failure modes: \{\textsc{Poor Enrollment}, \textsc{Safety/Adverse Effect}, \textsc{Drug Lack of Efficacy}\}, the Analyzer Agent produces a prioritized set of modifications based on Protocol Taxonomy and Action Determination, targeting the protocol feature under consideration; then specifies the action strategy \{\textsc{Delete}, \textsc{Modify}, \textsc{Add}\} (with \textsc{Keep} as the no-op default) and a confidence score.

\noindent\textbf{Protocol Taxonomy.}~ Before being processed through the agent redesign pipeline, protocol features are classified using medical terminology into structured categories: eligibility criteria, safety exclusions, selection criteria, and enrichment criteria. We further categorized dosage and outcomes by safety risk, failure contribution, and modification efficacy. These classifications guide action selection based on the observed failure reason and its corresponding analysis.

\noindent\textbf{Action Determination.}~ 
The agent assigns each candidate criterion a confidence score derived from its taxonomy classification and the observed failure mode. Two memory signals calibrate these scores: a cross-trial prior over edit-category effectiveness shifts confidence toward historically productive edit types (§\ref{sec:cross_trial}), while within-trial feedback from prior iterations penalizes already-explored or previously-failed criteria and rewards unexplored ones (§\ref{sec:within_trial}). A mechanism alignment check synthesizes protocols within the intervention's mechanism of action. The prioritized targets, balancing domain knowledge with empirical feedback, guide the Generator Agent (§\ref{sec:augment_agent}).

\subsubsection{Protocol Refinement Generator}
\label{sec:augment_agent}

The Generator Agent translates diagnostic insights from the Analyzer Agent into diverse design refinements that address identified weaknesses while preserving clinical validity.

\noindent\textbf{Type Conditioned Constrained Generation.}~
The agent employs action-specific logic: \textsc{Delete} critical failure factors while preserving safety; \textsc{Modify} adjusts thresholds or operationalizes vague terms; \textsc{Add} introduces biomarker enrichment or contraindication criteria. Various modifications are first validated for clinical safety (§\ref{sec:validation_agent}), then proceed to formal evaluation.

\subsubsection{Clinical Safety Validator}
\label{sec:validation_agent}

To ensure the proposed augmentations satisfy clinical safety standards, the system uses \textit{LLM-as-a-Judge} \citep{zheng2023judgingllmasajudgemtbenchchatbot} with domain database retrieval and autonomous validation.

\noindent\textbf{Database Retrieval.}~
The multi-agent system enhances embedded parametric knowledge with targeted 
retrieval from biomedical databases: DrugBank~\citep{wishart2018drugbank} with pharmacological profiles including toxicity, metabolism, contraindications; Disease Database~\citep{chen2024cod} that contains diagnostic criteria, symptomatology, risk factors; and PubMed Abstract spanning 1975-2025. Retrieval employs dense embeddings (BioBERT~\citep{lee2020biobert} for drugs/diseases, PubMedBERT~\citep{gu2021domain} for literature) with FAISS indexing. Retrieved results have tangential content filtered, while enforcing strict temporal constraints that limit PubMed queries to prevent outcome leakage.

\noindent\textbf{Autonomous Safety Validation.}~
The agent checks and prunes unsafe modifications (dosage changes, population shifts, contraindications) triggering retrieval from external databases when identifying knowledge gaps in medical assessments. Validated candidates that passed the validation proceed to the Evaluator (§\ref{sec:exploration_orchestrator}) for simulation-based evaluation. 

\subsubsection{Protocol Candidate Evaluator}
\label{sec:exploration_orchestrator}

The Evaluator combines validated augmentations with the original trial protocol into complete redesigned trial candidates, each evaluated through simulation-based assessment with a success probability assigned that guides modification and hierarchical learning (§\ref{sec:hierarchical_learning}).

\noindent\textbf{Search Strategy.}~
Candidate trials are formed by combining validated augmentations across original protocols, where Beam Search \cite{Freitag_2017} is used to reduce exponential complexity to approximately quadratic.

\noindent\textbf{Simulation Environment.}~
To provide reliable feedback for agent-generated modifications, we train a trial candidates' outcome predictor from encoded trial features (Appendix~\ref{app:ablation}), serving as a learned scalar verifier that enables rapid evaluation for thousands of redesigns without conducting actual clinical trials, guiding the agent system toward promising protocol optimization. The improvement in predicted success probability, from the original trial to the redesigned candidate, serves as the reward signal. While validated augmentations enter this stage, the reward is defined over a clinically admissible candidate space.

\subsection{In-Context Learning with Multi-level Memory}
\label{sec:hierarchical_learning}

Without memory or iterative feedback, multi-agent systems repeat failed strategies and cannot leverage historical performance. We address this via In-Context Learning \cite{brown2020language} from reward guided feedback (§\ref{sec:reward_model}) with knowledge consolidation operating at two temporal scales: \textit{within-trial learning} (§\ref{sec:within_trial}) accumulates local memory across iterations for trial refinement, while \textit{cross-trial learning} (§\ref{sec:cross_trial}) maintains global memory to transfer successful patterns across the trial corpus.

\subsubsection{Redesign Reward}
\label{sec:reward_model}

To identify specific modifications that drive improvement, we decompose protocol-level redesigns outcome probabilities $p(T')$ into augmentation-level rewards. The Evaluator first evaluates combined trial variants via prediction, then attributes credit to individual modifications. For each augmentation $m$, we compute its marginal contribution $r(m)$ across the explored combinatorial space:
\begin{equation}
\label{eq:marginal_reward}
r(m) = \mathbb{E}_{m\in T'}[p(T')] - \mathbb{E}_{m\not\in T' }[p(T')]. 
\end{equation} 
The complete reward distribution $\mathcal{R}_t$ encompasses all redesign augmentations with their validation status and rewards, enabling performance stratified knowledge extraction from both successful modifications and contraindicated patterns.

\subsubsection{Within-trial Learning: Local Iterative Optimization}
\label{sec:within_trial}

To refine LLM behavior by prompt optimization \cite{ramnath2025systematic}, we extract short-term memory from reward $\mathcal{R}_t$ and integrate with Agent.

\paragraph{Knowledge Extraction.}
We partition $\mathcal{R}_t$ into two types of knowledge: action-level patterns $\mathcal{K}^s_t$ aggregate modification type performance across aspects; and example-level demonstrations $\mathcal{K}^t_t$ comprise performance stratified modifications.


\paragraph{Agent Integration.} The Analyzer integrates $\mathcal{K}^s_t$ through coverage-based confidence adjustment over the prior iteration's outcomes: criteria whose augmentations yielded positive reward are down-weighted (discouraging redundant re-editing), modification strategies whose entire augmentation set failed validation or produced negative reward are excluded, and criteria left unexplored receive a confidence bonus to encourage coverage. To prevent the search from re-converging on the same aspects, a fraction of selection slots in later iterations is reserved for criteria not selected in any prior iteration. The Generator Agent samples performance stratified exemplars $\mathcal{K}^t_t$ for few-shot prompting~\cite{brown2020language}. The Evaluator maintains a redesign pool of positively rewarded modifications for combinatorial search reuse, enabling \textit{test-time search space scaling}~\cite{snell2024scaling}.



\subsubsection{Cross-trial Learning: Global Memory}
\label{sec:cross_trial}

After each trial converges, its reward distribution $\mathcal{R}_t$ is distilled via LLM synthesis into a global memory pool, so that every trial both benefits from and contributes to an evolving knowledge base~\cite{parisi2019continual}. For each Edit category (protocol-aspect-action) triple, the pool records the empirical success rate and mean reward over validated augmentations. This cross-trial prior is consumed at two stages:

\noindent\textbf{Confidence Calibration.}~ The Analyzer maps each candidate criterion to an edit category and shifts its confidence toward categories with high historical return, which optionally consider high-performing redesign patterns by upgrading a default \textsc{Keep} criterion to \textsc{Modify}. A bounded number of additional aspect slots is reserved for calibrated criteria that the base ranking would otherwise miss.

\noindent\textbf{Budget Allocation.}~ Under a fixed per-trial generation budget, the Generator distributes samples across aspects according to each category's historical return, starving categories with consistently low validated returns. Because reallocation operates at the category level --- not over specific augmentation texts --- and every augmentation passes the independent Safety Validator before reward attribution, this calibration cannot bypass clinical constraints to exploit predictor artifacts.
\section{Experiment}\label{sec:results}
We evaluate \mname\ across two dimensions: (1) simulation environment performance, validating that simulation environment achieves sufficient accuracy to serve as reliable feedback oracles, and (2) Agent optimization quality, demonstrating that our multi-agent system successfully redesigns failed trials through iterative learning.

\subsection{Experimental Setup}
Our evaluation consists of two parts. First, we evaluate the simulation environment at scale to verify that it provides a reliable feedback signal. Second, we evaluate whether \mname\ can effectively improve failed trial protocols through iterative redesign while producing protocol modifications that remain clinically plausible.

\noindent\textbf{Simulation Environment.}~
We use over 12000 annotated Phase I--IV trials from TrialBench~\citep{chen2025trialbench}. We encode multi-modal trial features into 6,173-dimensional embeddings and train LightGBM~\citep{ke2017lightgbm} classifiers to predict trial outcome. The detailed training process is deferred to Appendix~\ref{app:encode_detail}. We split the train-test 8:2 and further split the training set 8:2 for training and validation. 

Our simulation environment model is compared against \textit{baseline} approaches, including TrialBench~\citep{chen2025trialbench} and HINT~\citep{fu2022hint}, prior state-of-the-art systems operating on the original TrialBench feature space. Additionally, a Logistic Regression model is trained on the same encoded features (Appendix~\ref{app:encode_detail}) for reference.

\noindent\textbf{Multi-Agent Redesign.}~
We use GPT-4o-mini as our base model and evaluate the full redesign pipeline at the \emph{protocol level}. Specifically, we focus on a stratified set of 60 failed trials from the held-out test set, with 20 trials from each failure mode: enrollment, safety, and efficacy. The sample is selected to cover different failure modes and trial phases, with additional dataset statistics reported in Appendix~\ref{app:data_stats}. We use exhaustive search when the combinatorial space has fewer than $1{,}000$ candidates, and otherwise use beam search with width $k=8$. We measure agent effectiveness using predicted probability improvement, threshold achievement rate, and convergence efficiency, and compare \mname's performance with GPT-4o LLM and the ReAct agent framework. We further complement the quantitative evaluation with diverse case studies against \emph{real-world expert redesigns}, examining whether \mname's proposed changes are clinically plausible and consistent with real-world trial design decisions.


\subsection{Failure-specific Prediction}
Implemented in \mname\ Agent, the simulation environment must correctly predict specific failure outcomes. We train three independent models on our encoded features, each targeting one failure detection task against success. \Cref{tab:binary_all} reports comprehensive metrics across our models and baseline approaches trained for the task: Poor Enrollment, Safety/Adverse Effect and Lack of Efficacy prediction. Our model achieves PR-AUC $> 0.75$ across all failure modes, providing reliable discriminative feedback. All models achieve Failure Detection Rates of 69.5-74.0\% with task-specific prediction thresholds ($p \geq 0.6$ for enrollment, $p \geq 0.9$ for safety, $p \geq 0.85$ for efficacy). This ensures that predicted probability shifts $\Delta p > 0.03$ indicate improved trial designs (Appendix~\ref{app:data_stats}). Also,
we further validate the same model architecture against existing benchmarks on the TrialBench 4-class classification task (predicting Success, Enrollment Failure, Safety Failure, or Efficacy Failure)~\citep{chen2025trialbench}. \Cref{tab:multiclass_performance} shows that the base model used in our simulation environment outperformed the baselines (Appendix~\ref{app:trial_bench_prediction_task}).

\begin{table}[ht]
\centering
\caption{Performance comparison across three binary prediction tasks. We use ``*'' to denote tasks where \mname\ outperforms the best baseline significantly (\textit{i.e.}, using statistical hypothesis testing). 
}
\label{tab:binary_all}
\begin{adjustbox}{width=1\linewidth}
\begin{tabular}{@{}l l c c c@{}}
\toprule
\textbf{Task} & \textbf{Model} & \textbf{ROC-AUC} ($\uparrow$) & \textbf{PR-AUC} ($\uparrow$) & \textbf{Fail Detection} ($\uparrow$) \\
\midrule
\multirow{4}{*}{\textbf{Enrollment}}
& TrialBench    & 0.613 {\scriptsize $\pm$ 0.007} & 0.626 {\scriptsize $\pm$ 0.011} & 0.525 {\scriptsize $\pm$ 0.013} \\
& HINT          & 0.534 {\scriptsize $\pm$ 0.010} & 0.613 {\scriptsize $\pm$ 0.012} & 0.580 {\scriptsize $\pm$ 0.018} \\
& Logistic Reg. & 0.622 {\scriptsize $\pm$ 0.010} & 0.696 {\scriptsize $\pm$ 0.012} & 0.669 {\scriptsize $\pm$ 0.012} \\
& \mname        & \textbf{0.676} {\scriptsize $\pm$ 0.009}* & \textbf{0.754} {\scriptsize $\pm$ 0.010}* & \textbf{0.740} {\scriptsize $\pm$ 0.012}* \\
\midrule
\multirow{4}{*}{\textbf{Safety}}
& TrialBench    & 0.587 {\scriptsize $\pm$ 0.017} & 0.892 {\scriptsize $\pm$ 0.006} & 0.427 {\scriptsize $\pm$ 0.035} \\
& HINT          & 0.513 {\scriptsize $\pm$ 0.014} & 0.882 {\scriptsize $\pm$ 0.009} & 0.459 {\scriptsize $\pm$ 0.031} \\
& Logistic Reg. & 0.612 {\scriptsize $\pm$ 0.018} & 0.909 {\scriptsize $\pm$ 0.008} & 0.422 {\scriptsize $\pm$ 0.029} \\
& \mname        & \textbf{0.656} {\scriptsize $\pm$ 0.018}* & \textbf{0.925} {\scriptsize $\pm$ 0.007}* & \textbf{0.695} {\scriptsize $\pm$ 0.028}* \\
\midrule
\multirow{4}{*}{\textbf{Efficacy}}
& TrialBench    & 0.692 {\scriptsize $\pm$ 0.012} & 0.862 {\scriptsize $\pm$ 0.006} & 0.565 {\scriptsize $\pm$ 0.020} \\
& HINT          & 0.559 {\scriptsize $\pm$ 0.013} & 0.841 {\scriptsize $\pm$ 0.008} & 0.525 {\scriptsize $\pm$ 0.021} \\
& Logistic Reg. & 0.665 {\scriptsize $\pm$ 0.015} & 0.886 {\scriptsize $\pm$ 0.009} & 0.549 {\scriptsize $\pm$ 0.025} \\
& \mname        & \textbf{0.746} {\scriptsize $\pm$ 0.013}* & \textbf{0.914} {\scriptsize $\pm$ 0.007}* & \textbf{0.725} {\scriptsize $\pm$ 0.021}* \\
\bottomrule
\end{tabular}
\end{adjustbox}
\end{table}
\noindent\textbf{Feature Importance Analysis.}~
We studied feature importance analysis with SHAP~\citep{lundberg2017unified}. Consistent with our hypothesis, eligibility, drug-disease interaction features, and endpoint alignment features are most important for outcomes prediction (Appendix~\ref{app:simu_env_ablation_study}).

\subsection{Protocol Optimization}
\label{sec:agent_result}
Having confirmed the simulation environment's reliability, we evaluate \mname\ Agent's ability to redesign failed clinical trials. 



\begin{table}[ht]
\centering
\caption{Probability shift analysis by failure mode. Threshold reports the share of trials whose optimized $p_{\text{final}}$ reaches the task-specific success cutoff. $p_0/p_{\text{final}}$: initial/optimized success probabilities; IQR: 25th--75th percentile of $\Delta p$.}
\label{tab:combined_convergence_shift}
\begin{adjustbox}{width=\linewidth}
\begin{tabular}{@{}lcccc@{}}
\toprule
\textbf{Mode} & \textbf{Threshold} & \textbf{$p_0$ / $p_{\text{final}}$} & \textbf{$\Delta p$} & \textbf{IQR} \\
\midrule
Enrollment & 13/20 (65\%) & 0.506 / 0.609 & +0.103 & [+0.081, +0.131] \\
Safety & 5/20 (25\%) & 0.790 / 0.851 & +0.061 & [+0.030, +0.079] \\
Efficacy & 16/20 (80\%) & 0.820 / 0.878 & +0.058 & [+0.039, +0.070] \\
\midrule
\textbf{Overall} & \textbf{34/60 (56.7\%)} & \textbf{0.705 / 0.779} & \textbf{+0.074} & \textbf{[+0.033, +0.093]} \\
\bottomrule
\end{tabular}
\end{adjustbox}
\end{table}



\noindent\textbf{Convergence Analysis.}~
\Cref{tab:combined_convergence_shift} reports convergence and probability-shift statistics across all failure modes. In general, the Agent consistently improves the design of the protocols in all tasks and achieves an average success probability gain of $\Delta p=+0.074$, increasing the probability of predicted success from $0.705$ to $0.779$ in general. Enrollment shows the largest improvement, with $\Delta p=+0.103$ and 13/20 trials exceeding the predefined improvement threshold. Overall, tested trials achieve a 56.7\% threshold-passing rate, suggesting that \mname\ produces reliable improvements across heterogeneous failure modes. 


\begin{figure}[t]
    \centering
    \includegraphics[width=0.9\linewidth]{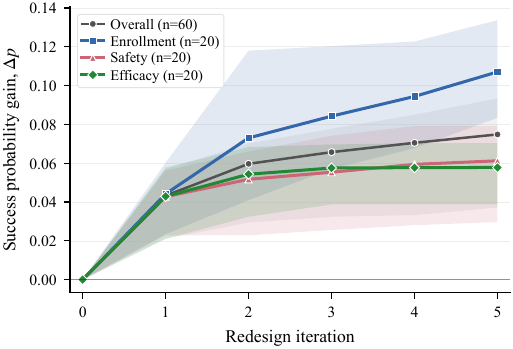}
    \vspace{-10pt}
    \caption{Iterative redesign trajectories by failure mode. Curves show mean cumulative improvement in predicted success probability ($\Delta p$).}
    \label{fig:probability_shift}
    \vspace{-10pt}
\end{figure}

\noindent\textbf{Iterative Improvement Trajectories.}~
We further examine the redesign trajectory over iterations. As shown in \Cref{fig:probability_shift}, the Agent steadily improves predicted success probability across iterations, reaching an overall mean gain of approximately $+0.074$ by iteration 5. Performance heterogeneity across failure modes reflects their different levels of amenability to protocol-level redesign. Enrollment failures achieve the largest gains, suggesting that recruitment-related issues can often be improved through eligibility and design refinements. Safety and efficacy also improve, but their gains are more constrained by drug-specific safety profiles and underlying therapeutic effectiveness.

Moreover, the learning trajectory reveals major initial gains followed by decreasing returns. This diminishing returns pattern validates the knowledge distillation mechanism: high-quality modifications are identified early through rapid retrieval boosted analysis, while later iterations exploit narrower optimization opportunities by refining secondary parameters or addressing edge case contraindications.


\noindent\textbf{Computational Cost and Efficiency.}~
The system demonstrates practical feasibility with a mean cost of \$0.156 per trial 
across 60 trials, a negligible fraction of the typical \$2.6 billion drug development cost, establishing \mname\ as practical for systematic protocol optimization at scale.
Full results on the computational cost are deferred to Appendix~\ref{app:cost}.


\begin{figure*}[ht]
    \centering
    \includegraphics[width=1\linewidth]{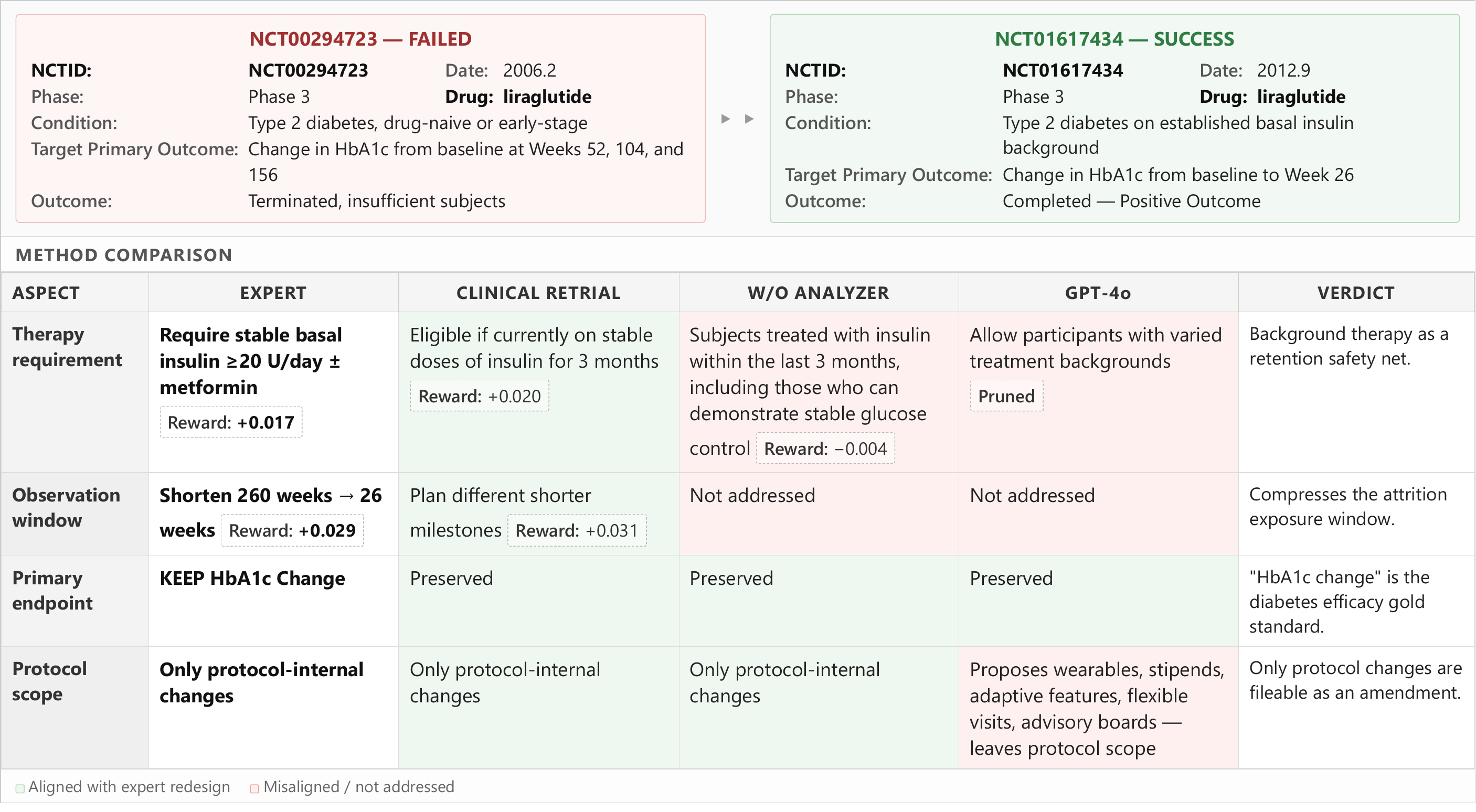}
    \vspace{-20pt}
    \caption{Case Study on one Poor Enrollment redesign: \textbf{(NCT00294723, 2006-02)}, and the real-world redesign \textbf{(NCT01617434, 2012-09)}. We demonstrate \mname's strategic alignment with the expert, comparing against w/o analyzer and zero-shot GPT-4o baseline.
    }
    \label{fig:enrollment_case}
    \vspace{-10pt}
\end{figure*}

\begin{table}[t]
\centering
\caption{Ablation study of the \mname\ framework across enrollment, safety, and efficacy failure modes. Each component is individually ablated to isolate its contribution. Reported values are pooled across the three tasks. The primary metric is Iter5 cumulative improvement, and significance is tested against the full system using one-sided paired $t$-tests with $H_1$: Full $>$ ablation. *$p < 0.05$, **$p < 0.01$, ***$p < 0.001$.}
\label{tab:agent_ablation}
\begin{adjustbox}{width=\linewidth,center}
\begin{tabular}{@{}lcccc@{}}
\toprule
\textbf{Method} & \textbf{N} & \textbf{Iter5 $\Delta p$} & \textbf{$p$-value} & \textbf{Cohen's $d_z$} \\
\midrule
Full System  & 60 & $+0.0741 {\scriptstyle \pm 0.048}$ & --- & --- \\
w/o Memory   & 60 & $+0.0532 {\scriptstyle \pm 0.044}$ & $(3.58{\times}10^{-8})^{***}$ & $0.79$ \\
w/o Pool     & 60 & $+0.0487 {\scriptstyle \pm 0.041}$ & $(3.82{\times}10^{-9})^{***}$ & $0.87$ \\
w/o Analysis & 60 & $+0.0522 {\scriptstyle \pm 0.039}$ & $(3.72{\times}10^{-8})^{***}$ & $0.79$ \\
GPT-4o & 60 & $+0.0265 {\scriptstyle \pm 0.019}$ & $(2.99{\times}10^{-19})^{***}$ & $1.68$ \\
ReAct & 60 & $+0.0465 {\scriptstyle \pm 0.039}$ & $(4.81{\times}10^{-9})^{***}$ & $0.86$ \\
\bottomrule
\end{tabular}%
\end{adjustbox}
\vspace{-7pt}
\end{table}

\noindent\textbf{Ablation Study on Self-improvement.}~
To validate the contribution of each architectural component, we conduct paired ablation studies across enrollment, safety, and efficacy failure modes. For each ablation, we remove one component while keeping the rest of the pipeline unchanged, and evaluate Iter5 cumulative improvement as the primary metric. Results pooled across all three tasks are reported in \Cref{tab:agent_ablation}, with task-specific ablation results provided in \Cref{sec:full_ablation} (\Cref{tab:agent_ablation_by_task}).

The full system achieves the highest Iter5 improvement ($\Delta p=+0.0741$), while all ablations significantly degrade performance. Removing the redesign pool causes the largest drop ($\Delta p=+0.0487$, $p=3.824{\times}10^{-9}$), followed by removing analysis ($\Delta p=+0.0522$, $p=3.723{\times}10^{-8}$) and memory ($\Delta p=+0.0532$, $p=3.582{\times}10^{-8}$). These results show that pool-based exploitation, structured diagnosis, and cross-trial memory all effectively contribute to the self-improvement capability of \mname.

\subsection{Case Studies}
\label{sec:case_study}



\noindent\textbf{Alignment with Real-world Trial Redesigns.}
We analyze diverse clinical trial pairs, where investigators successfully redesigned and re-executed failed protocols spanning enrollment, safety, and efficacy failure modes against \mname's redesign to validate its applicability, as well as to provide insight into clinical applicability. Such trial pair cases are chosen based on the mechanistic interpretability for systematic alignment measures.

In \Cref{fig:enrollment_case}, we present a poor enrollment redesign case in which the same sponsor (Novo Nordisk's Victoza\textsuperscript{\textregistered}) re-executed a failed design and achieved a positive readout. The study drug, \textit{liraglutide}, is a foundational GLP-1 agonist for type 2 diabetes (T2D). \mname~is compared against the expert~\cite{ahmann2015efficacy}, an ablation without the Failure Analyzer (w/o~Analyzer), and zero-shot GPT-4o. We defer the remaining case studies to Appendix~\ref{sec:case_study_detail}. As shown in \Cref{fig:enrollment_case}, \mname{} achieves alignment with
the expert: it matches the direction of primary fixes, preserves the criterion the expert deliberately leaves intact, while staying within
the regulatory amendment scope. The scalar verifier reliably separates direction-aligned from direction-opposed actions by reward sign, while the upstream validator prunes operations that would compromise scientific safety.

\noindent\textbf{RAG-Grounded Validator Case Studies.}~
We further present three validator case studies to highlight the role of RAG in improving the clinical reliability of \mname. Across these examples, disease-specific, literature-based, and drug-level retrieval provides external medical evidence that enables the Validator Agent to identify unsafe or misaligned redesigns that would otherwise appear plausible at the language level. 
Full case descriptions are deferred to Appendix~\ref{app:validator_cases}.


\section{Conclusion}\label{sec:conclusion}

In this work, we propose \mname, a self-evolving agent that moves beyond passive clinical trial outcome prediction to enable proactive optimization of clinical trial protocols. Evaluated on the standardized TrialBench benchmark, \mname\ achieves strong predictive performance while demonstrating the ability to discover significant protocol improvements with clinical best practices, highlighting the potential of agentic AI systems to serve as practical clinical decision support systems for more efficient autonomous trial design. Extensive case studies on trial pairs validate that \mname\ exhibits strong strategic alignment with independently derived real-world trial modifications, suggesting generalizable principles for protocol redesign. 



\section*{Limitations}
\label{sec:limitations}

Our framework has several limitations that suggest directions for future research. First, the simulation environment's predictive accuracy creates potential for false improvement signals and missed opportunities, though validation filtering and iterative refinement provide partial mitigation; this constraint could be addressed by integrating future state-of-the-art prediction models as drop-in replacements. Second, retrospective case study analysis reveals tactical domain knowledge gaps: while the system excels at strategic-level reasoning, it may struggle with operational specifics such as selecting appropriate biomarkers, anticipating implementation constraints, and distinguishing when radical simplification outperforms incremental fortification, often over-relying on complexity multiplication where parsimony proves more effective.

\section*{Acknowledgments}

Tianfan Fu is supported by the Young Scientists Fund (C Class) of the National Natural Science Foundation of China (Grant No.~62506154), the Fundamental Research Funds for the Central Universities, the Nanjing University International Collaboration Initiative (Grant No.~020214380129), and the ``111 Center'' (No.~B26023).

This work was partially supported by fundings from Coefficient Giving. We also appreciate the support from the Foundation Models and Applications Lab and ND-IBM Tech Ethics Lab at Notre Dame.




\bibliography{colm2026_conference}

\clearpage
\onecolumn
\appendix

\crefalias{section}{appendix}

\startcontents[appendix]
\part*{Appendix}
\printcontents[appendix]{}{1}{}

\clearpage
\section{\mname\  Algorithm}
\label{app:algo}
\begin{algorithm*}[ht]
\caption{Clinical trial optimization with In-Context Learning and Multi-level Memory.}
\label{alg:main}
\begin{algorithmic}[1]
\REQUIRE Failed trial $T_0$, failure mode $y \in \{\text{enrollment, safety, efficacy}\}$, global memory $\mathcal{M}^{\text{global}}$ \hfill \ENSURE Optimized protocol $T^*$, best reward $r_{\text{best}}$
\STATE Initialize: $r_{\text{best}} \leftarrow 0$, $T^* \leftarrow T_0$, $\mathcal{M}_0^{\text{local}} \leftarrow \emptyset$, $\mathcal{H}_0 \leftarrow \emptyset$
\FOR{$t = 1$ to $N_{\text{max}}$}
    \STATE $\mathcal{K}^s_t, \mathcal{K}^t_t \leftarrow$ \textbf{LoadMemory}$(\mathcal{M}^{\text{global}}[y], t)$ 
    \STATE $\mathcal{S}_t \leftarrow$ \textbf{AnalyzerAgent}$(T_{t-1}, y, \mathcal{M}_{t-1}^{\text{local}}, \mathcal{K}^s_t)$
    \STATE $\mathcal{A}_t \leftarrow$ 
    \textbf{GeneratorAgent}$(\mathcal{S}_t, T_{t-1}, \mathcal{M}_{t-1}^{\text{local}}, \mathcal{K}^t_t)$
    \STATE $\mathcal{R}_t, r_{\text{max}} \leftarrow$ \textbf{ExploreSearch}$(\mathcal{A}_t, \mathcal{H}_{t-1}, T_{t-1})$ 
    \STATE \textbf{if} $r_{\text{max}} > r_{\text{best}}$ \textbf{then} $r_{\text{best}} \leftarrow r_{\text{max}}$, $T^* \leftarrow \arg\max_{T' \in \mathbb{T}} f_{\theta}(T')$
    \STATE $\mathcal{K}_t \leftarrow$ \textbf{DistillKnowledge}$(\mathcal{R}_t)$; $\mathcal{H}_t \leftarrow \mathcal{H}_{t-1} \cup$ \textbf{ExtractPool}$(\mathcal{R}_t)$         
    \STATE $\mathcal{M}_t^{\text{local}} \leftarrow \mathcal{M}_{t-1}^{\text{local}} \cup \{\mathcal{K}_t, \mathcal{R}_t, \mathcal{H}_t\}$
\ENDFOR
\STATE $\mathcal{M}^{\text{global}} \leftarrow \mathcal{M}^{\text{global}} \cup$ \textbf{TransferMemory}$(T^*, \mathcal{M}_{N_{\text{max}}}^{\text{local}})$; \quad \RETURN $T^*$, $r_{\text{best}}$
\end{algorithmic}
\end{algorithm*}


\section{Simulation Environment Details}\label{app:ablation}

\subsection{Dataset Statistics}\label{app:data_stats}

Dataset used for training the prediction models comprises 20,769 clinical trials from TrialBench's failure reason dataset. Table~\ref{tab:data_distribution} shows the label distribution across four categories.

\begin{table}[h]
\centering
\caption{Distribution of failure reason labels in the dataset.}
\label{tab:data_distribution}
\begin{tabular}{@{}lrr@{}}
\toprule
\textbf{Failure Reason} & \textbf{Count} & \textbf{Percentage} \\
\midrule
Success & 9,939 & 47.8\% \\
Poor Enrollment & 7,229 & 34.8\% \\
Inefficacy & 2,217 & 10.7\% \\
Adverse Effect & 1,384 & 6.7\% \\
\midrule
\textbf{Total} & \textbf{20,769} & \textbf{100.0\%} \\
\bottomrule
\end{tabular}
\end{table}

The class imbalance reflects real-world trial outcomes: enrollment challenges are the most common failure mode (34.8\%), followed by efficacy gaps (10.7\%), while safety failures are relatively rare (6.7\%) due to rigorous preclinical screening. Success cases (47.8\%) include trials that completed without major protocol violations or early termination.

Due to computational cost constraints, we randomly select a stratified sample of 60 trials from the test set (20 enrollment, 20 safety, 20 efficacy), ensuring representation across failure modes and trial phases. Table~\ref{tab:dataset_phase} presents the phase composition.

\begin{table}[h]
\centering
\caption{Test distribution by trial phase.}
\label{tab:dataset_phase}
\begin{tabular}{lcc}
\toprule
\textbf{Phase} & \textbf{Count} & \textbf{\%} \\
\midrule
Phase 1 & 8 & 13.3 \\
Phase 2 & 27 & 45.0 \\
Phase 3 & 15 & 25.0 \\
Phase 4 & 10 & 16.7 \\
\midrule
{Total} & {60} & {100} \\
\bottomrule
\end{tabular}
\end{table}

\subsection{Encode Details}\label{app:encode_detail}

This appendix provides comprehensive implementation details for the Simulation Environment described in §\ref{sec:exploration_orchestrator}, including encoder pretraining procedures, model training hyperparameters, and detailed validation results.

\paragraph{Text Features.} Textual contents are encoded using BioBERT~\citep{lee2020biobert}, a domain-adapted language model pre-trained on PubMed abstracts and PMC full-text articles. Critically, we diverge from prior work by decomposing eligibility criteria at the sentence level rather than treating them as monolithic text blocks. For each text field $\mathcal{T}$, we decompose it into sentences $\mathcal{T} = \{s_1, s_2, \ldots, s_n\}$. Each sentence is encoded via BioBERT and the final text embedding is obtained via max pooling:
\begin{equation}
\label{eq:biobert_encoding}
\mathbf{e}_{s_i} = \text{BioBERT}(s_i), \quad \mathbf{h}_{\mathcal{T}} = \max_{i=1}^{n} \mathbf{e}_{s_i}
\end{equation}
This sentence-level representation preserves granularity essential for aspect-specific modification: \mname\ Agent can target individual criteria rather than generic protocol summaries.

\paragraph{Graph Features.} We incorporate pre-trained molecular and disease encodings to capture pharmacological properties and disease characteristics.

\textit{Drug Molecular Graphs.} Each drug molecule $m$ is represented as a graph $\mathcal{G}_m = (\mathcal{V}, \mathcal{E})$ where nodes $v \in \mathcal{V}$ are atoms and edges $(u,v) \in \mathcal{E}$ are bonds. We employ Message Passing Neural Networks (MPNNs) to aggregate neighborhood information over $L$ iterations:
\begin{equation}
\label{eq:mpnn_message}
\mathbf{m}_{uv}^{(l)} = \text{ReLU}\left(W_i \cdot [\mathbf{f}_u \oplus \mathbf{f}_{uv}] + W_h \cdot \sum_{w \in \mathcal{N}(u) \setminus v} \mathbf{m}_{wu}^{(l-1)}\right), 
\end{equation}
where $\mathbf{m}_{uv}^{(l)} \in \mathbb{R}^{d_{\text{mpnn}}}$ is the message from atom $u$ to atom $v$ at layer $l$, $\mathcal{N}(u)$ denotes neighbors of $u$, $\oplus$ denotes concatenation, and $W_i, W_h$ are learnable transformation matrices. After $L$ message passing iterations, node embeddings are computed as:
\begin{equation}
\label{eq:mpnn_node}
\mathbf{h}_u = \text{ReLU}\left(W_o \cdot \left[\mathbf{f}_u \oplus \sum_{v \in \mathcal{N}(u)} \mathbf{m}_{vu}^{(L)}\right]\right). 
\end{equation}
The graph-level drug embedding is obtained via global average pooling:
\begin{equation}
\label{eq:mpnn_readout}
\mathbf{h}_{\text{drug}} = \frac{1}{|\mathcal{V}|} \sum_{u \in \mathcal{V}} \mathbf{h}_u \in \mathbb{R}^{d_{\text{mpnn}}}
\end{equation}
For trials with multiple drugs, we average their embeddings. The MPNN encoder is pretrained on pharmacokinetic (ADMET) tasks, then fine-tuned on trial outcome labels (details in Appendix~\ref{app:ablation}).

\textit{Disease Hierarchical Encoding.} Each disease is represented by an ICD-10 code $d_i$ following a hierarchical taxonomy with ancestors $\mathcal{A}(d_i) = \{a_1, a_2, \ldots, a_p\}$. We use Graph-based Attention Model (GRAM) to encode hierarchical disease information. Each code $c$ has a learnable base embedding $\mathbf{e}_c \in \mathbb{R}^{d_{\text{gram}}}$. The hierarchical embedding for disease $d_i$ is computed as an attention-weighted sum over itself and its ancestors:
\begin{equation}
\label{eq:gram_hierarchical}
\mathbf{h}_{d_i} = \sum_{a_j \in \mathcal{A}(d_i) \cup \{d_i\}} \alpha_{ji} \cdot \mathbf{e}_{a_j}
\end{equation}
where the attention weight $\alpha_{ji}$ measures the relevance of ancestor $a_j$ to the current disease $d_i$:
\begin{equation}
\label{eq:gram_attention}
\alpha_{ji} = \frac{\exp(\phi([\mathbf{e}_{a_j} \oplus \mathbf{e}_{d_i}]))}{\sum_{a_k \in \mathcal{A}(d_i) \cup \{d_i\}} \exp(\phi([\mathbf{e}_{a_k} \oplus \mathbf{e}_{d_i}]))}
\end{equation}
where $\phi(\cdot): \mathbb{R}^{2d_{\text{gram}}} \rightarrow \mathbb{R}$ is a learnable single-layer network. For trials targeting multiple diseases, we average their embeddings. The GRAM encoder is initialized with the ICD-10 hierarchical ontology, then fine-tuned on historical trial success rates (details in Appendix~\ref{app:ablation}).

\paragraph{Tabular Features.} We encode structured trial metadata through a modular pipeline that processes categorical attributes, demographic constraints, administrative properties, and enrollment characteristics. The pipeline extracts 29 numerical features.

\noindent\textbf{Problem Formulation and Dataset.} We formulate clinical trial outcome prediction as a binary classification problem over three distinct failure modes: poor enrollment, safety/drug adverse effect, and drug inefficacy. We train models separately for each failure mode, enabling \mname\ Agent to target specific causes during protocol optimization. Our experiments utilize the TrialBench dataset~\citep{chen2025trialbench}, which contains over 12,000 annotated clinical trials spanning Phase I through Phase IV, with each trial labeled according to outcome. The dataset provides multi-modal features including drug molecular structures, disease ICD-10 codes, eligibility criteria text, trial metadata, and intervention details. Following standard practice to avoid temporal leakage, we partition data chronologically by trial completion year, 8:2 train vs test. According to Table~\ref{tab:features}, features are encoded into total dim=6,173.

\begin{table}[H]
\centering
\caption{Feature specification summary. Novel contributions include sentence-level eligibility parsing and fine-tuned molecular-disease encoders.}
\label{tab:features}
\small
\begin{tabular}{@{}llrl@{}}
\toprule
\textbf{Category} & \textbf{Component} & \textbf{Dim} & \textbf{Method} \\
\midrule
\multirow{6}{*}{Text} 
& Study Design & 768 & BioBERT \\
& Dosage & 768 & BioBERT \\
& Intervention & 768 & BioBERT + pooling \\
& Condition & 768 & BioBERT + pooling \\
& Eligibility Inclusion & 768 & BioBERT + pooling \\
& Eligibility Exclusion & 768 & BioBERT + pooling \\
\midrule
\multirow{2}{*}{Graph} 
& Drug (ADMET) & 768 & MPNN (fine-tuned) \\
& Disease (ICD) & 768 & GRAM (fine-tuned) \\
\midrule
\multirow{4}{*}{Tabular} 
& Categorical Features & 18 & One-Hot \\
& Age constraints & 2 & Unit normalization \\
& Multi-hot indicators & 9 & Binary encoding \\
\midrule
\multicolumn{2}{l}{\textbf{Total}} & \textbf{6,173} & \\
\bottomrule
\end{tabular}
\end{table}

\textbf{Feature Concatenation and Prediction.} All feature modalities are concatenated into a single input vector:

\begin{equation}
\label{eq:gbdt_input}
\mathbf{x}_{\text{trial}} = [\mathbf{h}_{\text{design}};\mathbf{h}_{\text{dose}}; \mathbf{h}_{\text{interv}}; \mathbf{h}_{\text{cond}};  \mathbf{h}_{\text{incl}}; \mathbf{h}_{\text{excl}}; \mathbf{h}_{\text{drug}}; \mathbf{h}_{\text{disease}}; \mathbf{f}_{\text{tabular}}] \in \mathbb{R}^{6173}
\end{equation}

where semicolons denote concatenation. For each failure mode $\tau \in \{\text{enrollment, safety, efficacy}\}$, we train a separate LightGBM classifier $\mathcal{M}_{\tau}$ that predicts trial success probability. The predicted probability $\hat{y} = \mathcal{M}_{\tau}(\mathbf{x}_{\text{trial}}) \in [0, 1]$ serves as the reward signal for evaluating protocol modifications in the agent system.

\textbf{Model Training and Validation.} We employ LightGBM~\citep{ke2017lightgbm} for its computational efficiency with high-dimensional sparse features. Three independent models are trained for enrollment, safety, and efficacy failure prediction using cross-validation with early stopping. The trained GBDT models achieve strong predictive performance across all failure modes (PR-AUC $>$ 0.75) with well-calibrated probability estimates, validating the simulation environment as a reliable proxy for real trial outcomes.

\subsection{Trial Bench Prediction Task}\label{app:trial_bench_prediction_task}

\noindent\textbf{TrialBench Benchmark.}~
We further validate the same model architecture against existing benchmarks on the TrialBench 4-class classification task (predicting Success, Enrollment Failure, Safety Failure, or Efficacy Failure)~\citep{chen2025trialbench}. \Cref{tab:multiclass_performance} shows that the base model used in our simulation environment outperformed most baselines, achieving a ROC-AUC improvement of $6\%$ to $19\%$.

\begin{table*}[ht]
\centering
\caption{Performance of multi-class clinical trial outcome prediction across trial phases.}
\label{tab:multiclass_performance}
\begin{adjustbox}{width=1.0\linewidth}
\begin{tabular}{@{}lcccccccc@{}}
\toprule
\multirow{2}{*}{\textbf{Model}} 
& \multicolumn{2}{c}{\textbf{Phase 1}} 
& \multicolumn{2}{c}{\textbf{Phase 2}} 
& \multicolumn{2}{c}{\textbf{Phase 3}} 
& \multicolumn{2}{c}{\textbf{Phase 4}} \\
\cmidrule(lr){2-3} \cmidrule(lr){4-5} \cmidrule(lr){6-7} \cmidrule(lr){8-9}
& \textbf{ROC-AUC} ($\uparrow$) & \textbf{PR-AUC} ($\uparrow$) 
& \textbf{ROC-AUC} ($\uparrow$) & \textbf{PR-AUC} ($\uparrow$) 
& \textbf{ROC-AUC} ($\uparrow$) & \textbf{PR-AUC} ($\uparrow$) 
& \textbf{ROC-AUC} ($\uparrow$) & \textbf{PR-AUC} ($\uparrow$) \\
\midrule
\textbf{TrialBench}
& 0.475 $\pm$ 0.027 & 0.255 $\pm$ 0.006 
& 0.569 $\pm$ 0.010 & 0.295 $\pm$ 0.008 
& 0.550 $\pm$ 0.012 & 0.279 $\pm$ 0.008 
& 0.477 $\pm$ 0.022 & 0.256 $\pm$ 0.007 \\
\textbf{HINT}
& 0.540 $\pm$ 0.022 & 0.272 $\pm$ 0.009 
& 0.535 $\pm$ 0.009 & 0.267 $\pm$ 0.005 
& 0.474 $\pm$ 0.019 & 0.251 $\pm$ 0.006 
& 0.548 $\pm$ 0.021 & 0.273 $\pm$ 0.014 \\
\textbf{Logistic Reg.}
& 0.606 $\pm$ 0.019 & 0.326 $\pm$ 0.015 
& 0.583 $\pm$ 0.011 & 0.306 $\pm$ 0.008 
& 0.621 $\pm$ 0.016 & 0.350 $\pm$ 0.017 
& \textbf{0.550 $\pm$ 0.022} & 0.280 $\pm$ 0.010 \\
\midrule
\textbf{\mname}
& \textbf{0.633 $\pm$ 0.016} & \textbf{0.344 $\pm$ 0.016} 
& \textbf{0.662 $\pm$ 0.011} & \textbf{0.382 $\pm$ 0.011} 
& \textbf{0.669 $\pm$ 0.017} & \textbf{0.412 $\pm$ 0.019} 
& 0.543 $\pm$ 0.025 & \textbf{0.282 $\pm$ 0.012} \\
\bottomrule
\end{tabular}
\end{adjustbox}
\end{table*}

\subsection{Ablation Study}\label{app:simu_env_ablation_study}

\textbf{Word-Level Attention Analysis.} Figure \ref{fig:sentence_importance} demonstrates the word-level attention weights captured by BioBERT embeddings, visualized through Shapley values. The heatmap reveals that clinical keywords such as ``woman,'' ``contraception,'' receive the highest attention weights (0.0208--0.0274), while functional words like prepositions and conjunctions are assigned lower weights. This attention distribution indicates that the model effectively focuses on medically relevant terminology when processing eligibility criteria, suggesting that domain specific language models can automatically identify critical phrases without explicit feature engineering.

\begin{figure}[H]
\centering



\begin{tcolorbox}[colback=gray!5, sharp corners, boxrule=0.5pt, boxsep=1mm, left=1mm, right=1mm, top=1mm, bottom=1mm, center title]
\begin{tabular}{@{}*{6}{c}@{}}
\colorbox{red!10}{-} &
\colorbox{red!30}{potentially} &
\colorbox{red!60}{fertile} &
\colorbox{red!80}{woman} &
\colorbox{red!10}{without} &
\colorbox{red!60}{$\beta$-hcg} \\
0.0001 & 0.0075 & 0.0150 & 0.0208 & 0.0001 & 0.0165 \\[2mm]
\colorbox{red!60}{negative} &
\colorbox{red!60}{harvested} &
\colorbox{red!30}{until} &
\colorbox{red!30}{48} &
\colorbox{red!60}{hours} &
\colorbox{red!30}{before} \\
0.0179 & 0.0175 & 0.0144 & 0.0142 & 0.0157 & 0.0153 \\[2mm]
\colorbox{red!60}{operation} &
\colorbox{red!60}{or} &
\colorbox{red!60}{not} &
\colorbox{red!60}{using} &
\colorbox{red!60}{acceptable} &
\colorbox{red!80}{contraception} \\
0.0184 & 0.0177 & 0.0168 & 0.0163 & 0.0156 & 0.0212 \\[2mm]
\colorbox{red!80}{for} &
\colorbox{red!60}{participation} &
\colorbox{red!10}{in} &
\colorbox{red!60}{this} &
\colorbox{red!80}{study} &
\\
0.0245 & 0.0166 & 0.0088 & 0.0155 & 0.0274 &
\end{tabular}
\end{tcolorbox}

\caption{
Visualization of text segments in the BioBERT encoder's output, illustrating Shapley values derived from Clinical Trials. Shapley values correspond to attention weights, with darker colors indicating higher weights.
}
\label{fig:sentence_importance}
\end{figure}

\textbf{Sentence-Level Eligibility Weights.} Table \ref{tab:sentence_level} illustrates a example of sentence-level importance scores within the inclusion criteria for trial NCT01102504, normalized across all eligibility statements, with weighted importance calculated on predict probability shift if masking out each eligibility protocols. The model assigns highest weights (0.20--0.25) to sentences describing acute cerebrovascular events such as ``Transient ischemic attack (TIA)'' and ``Stroke (ipsilaterally to the stenotic artery),'' while demographic criteria like age receive minimal attention (0.07). Notably, the quantitative stenosis threshold ``$>30\%$ stenosis on initial B-mode ultrasonography imaging'' receives substantial weight (0.18), indicating that the model prioritizes disease severity markers and clinical events over basic demographic qualifications when predicting trial outcomes.

\begin{table}[h!]
\centering
\caption{Inclusion Criteria with Sentence Importance (Color-coded)}
\begin{tabular}{p{11cm} c}
\toprule
\textbf{NCT01102504 Eligibility Criteria Protocols} & \textbf{Weight} \\
\midrule

\textbf{\textcolor{gray}{Inclusion Criteria:}} & \textcolor{gray}{0.03} \\[6pt]

\hspace{1em}- \textcolor{gray}{Age 40--90 years old,} & \textcolor{gray}{0.07} \\[6pt]

\hspace{1em}- Clinically documented carotid symptomatic atherosclerotic disease (symptomatic disease will be considered if one of the following has occurred within 2 months prior to symptoms:) & 0.12 \\[6pt]

\hspace{3em}\textcolor{black}{1. Amaurosis fugax} & 0.10 \\[4pt]
\hspace{3em}\textcolor{red}{2. Transient ischemic attack (TIA)} & \textcolor{red}{0.20} \\[4pt]
\hspace{3em}\textcolor{red}{3. Stroke (ipsilaterally to the stenotic artery)} & \textcolor{red}{0.25} \\[6pt]

\hspace{1em}\textcolor{red}{- $>30\%$ stenosis on initial B-mode ultrasonography imaging,} & \textcolor{red}{0.18} \\[6pt]

\hspace{1em}- \textcolor{gray}{Written, informed consent.} & \textcolor{gray}{0.05} \\

\bottomrule
\end{tabular}
\label{tab:sentence_level}
\end{table}

\begin{figure}[h]
\centering
\includegraphics[width=1\linewidth]{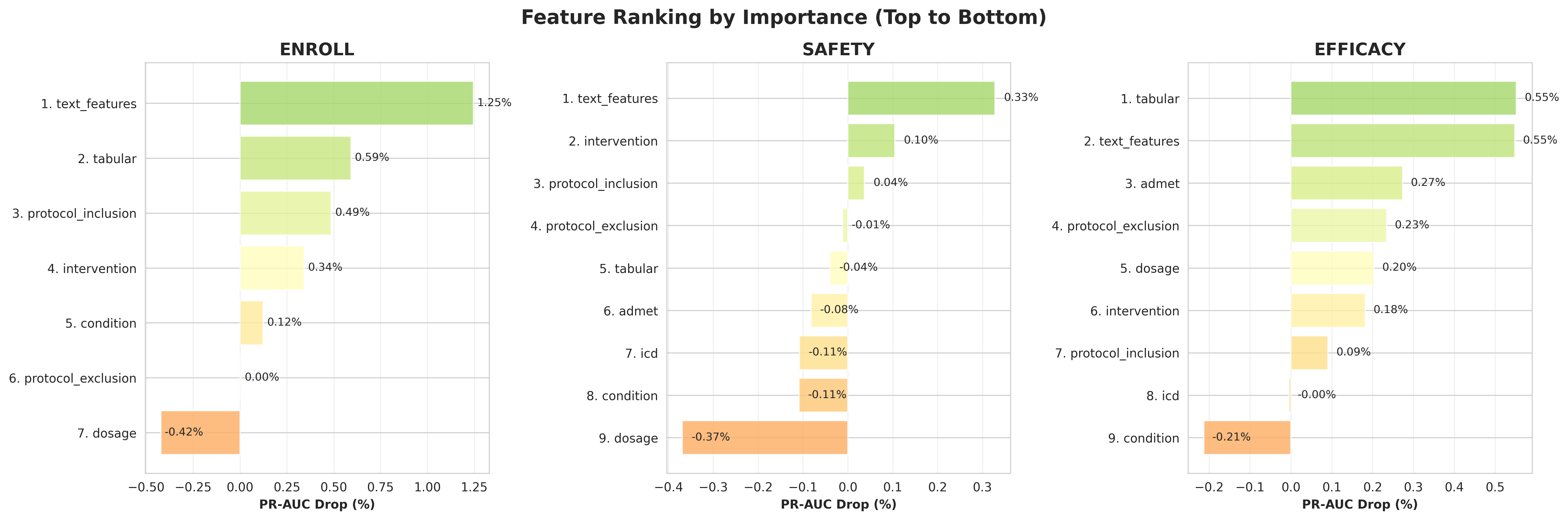}
\caption{Feature importance of each encoder on 3 classification tasks (enrollment, safety, and efficacy), measured by PR-AUC drop when the encoder is masked out during prediction.}
\label{fig:feature_importance}
\end{figure}

\textbf{Encodes Contributions Revealed Through Ablation Analysis.} Figure \ref{fig:feature_importance} presents the relative importance of different encoders across three prediction tasks through systematic masking experiments. By individually masking each encoder and measuring the resulting PR-AUC drop, we quantify each component's contribution to enrollment, safety, and efficacy outcome predictions. The analysis reveals task-specific dependency patterns: certain encoders prove critical for particular outcomes, with their removal causing substantial performance degradation, while showing minimal impact on other tasks. This heterogeneous importance distribution demonstrates that different aspects of trial design and patient characteristics drive distinct clinical endpoints. The varying magnitudes of PR-AUC drops across tasks validate the multi-task learning framework's ability to capture task-specific representations while identifying which shared features are most crucial for each prediction objective.

\section{Deferred Self-Evolving Multi-Agent Experiments}
\subsection{Computation Cost}
\label{app:cost}
\Cref{tab:cost_table} summarizes the per-trial computational cost of \mname\ across the three failure modes. Overall, the optimization process costs only \$0.156 per trial on average, indicating that the proposed self-evolving search procedure remains inexpensive even when applied across diverse protocol redesign tasks. The cost-efficiency ratio is also stable across modes, with an overall cost of \$2.11 per unit improvement in predicted success probability.

\begin{table}[h]
\centering
\caption{Computational cost and efficiency across failure modes. Cost and $\Delta p$ report per-trial averages; Cost/$\Delta p$ measures dollars spent per unit improvement in predicted success probability.}
\label{tab:cost_table}
\begin{tabular}{lcccc}
\toprule
\textbf{Mode} & \textbf{N} & \textbf{Cost (\$)} & \textbf{$\Delta p$} & \textbf{Cost/$\Delta p$} \\
\midrule
Enrollment & 20 & 0.247 & +0.103 & 2.40 \\
Safety     & 20 & 0.144 & +0.061 & 2.35 \\
Efficacy   & 20 & 0.078 & +0.058 & 1.34 \\
\midrule
\textbf{Overall} & \textbf{60} & \textbf{0.156} & \textbf{+0.074} & \textbf{2.11} \\
\bottomrule
\end{tabular}
\end{table}

\subsection{Cross-Predictor Evaluation}
\label{app:cross_predictor}

A central concern for any surrogate-driven optimizer is whether the reported
gains reflect genuine protocol improvement or merely the exploitation of
artifacts specific to the reward model. Under this concern, we conduct a
cross-predictor evaluation using HINT~\citep{fu2022hint} as an independent
verifier. HINT is never used during agent optimization, reward computation,
candidate selection, or prompt adaptation. We freeze the redesigns generated under the Simulation Environment reward and re-score both the original and the final protocol of every trial with HINT. Table~\ref{tab:cross_predictor} reports the resulting HINT-measured gains alongside the Simulation Environment (GBDT) reference gains from Table~\ref{tab:cost_table}; per-trial variability is summarized by the interquartile range (IQR) of $\Delta p$.

\begin{table}[h]
\centering
\caption{Cross-predictor validation of agent gains. Redesigns optimized under
the LightGBM reward model are re-scored with an independent HINT verifier.
$p_0$ and $p_{\text{final}}$ are HINT probabilities for the original and final
protocols; IQR is the interquartile range of the per-trial HINT $\Delta p$; the
last column repeats the GBDT reference $\Delta p$ from Table~\ref{tab:cost_table}.}
\label{tab:cross_predictor}
\small
\begin{tabular}{@{}lccccc@{}}
\toprule
\textbf{Mode} & \textbf{$p_0$} & \textbf{$p_{\text{final}}$} & \textbf{$\Delta p$} & \textbf{IQR} & \textbf{GBDT ref $\Delta p$} \\
\midrule
Enrollment & 0.539 & 0.645 & $+0.106$ & $[+0.047, +0.148]$ & $+0.103$ \\
Safety     & 0.862 & 0.878 & $+0.017$ & $[+0.008, +0.022]$ & $+0.061$ \\
Efficacy   & 0.334 & 0.415 & $+0.081$ & $[+0.004, +0.118]$ & $+0.058$ \\
\midrule
\textbf{Overall} & \textbf{0.578} & \textbf{0.646} & \textbf{$+0.068$} & \textbf{$[+0.010, +0.112]$} & \textbf{$+0.074$} \\
\bottomrule
\end{tabular}
\end{table}

\paragraph{The gains survive an independent re-scoring.} Under HINT, the agent's
redesigns yield a positive $\Delta p$ in every failure mode, and the IQR lower
bound stays above zero in all rows, indicating that the improvement is not
carried by a few outliers. The two predictors are trained independently and
disagree substantially on the \emph{absolute} baseline probability
(e.g., HINT rates the failed safety protocols at $p_0 = 0.862$), yet they agree
on the \emph{direction} of the change. Enrollment and the overall gain match the
GBDT reference closely ($+0.106$ vs.\ $+0.103$; $+0.068$ vs.\ $+0.074$). The
safety gain is attenuated ($+0.017$ vs.\ $+0.061$), which reflects a ceiling
effect (HINT already scores these protocols near the top of its range), together
with the drug-intrinsic difficulty of safety fixes, rather than a reversal of
sign.

\paragraph{Scope of the evidence.} We read this result as directional
corroboration rather than absolute confirmation. HINT is an independent but
imperfect model: its weaker absolute calibration explains the baseline gap
above, which is why the cross-check relies on the sign and rough magnitude of
$\Delta p$ rather than on absolute probability levels. HINT and our reward model
also consume the same input modalities (eligibility text, molecular graphs,
disease codes), so the two are independent \emph{models} rather than fully
independent \emph{representations}. With those caveats, the agreement across an
independently trained predictor materially reduces the likelihood that the
reported gains are driven by LightGBM-specific artifacts. It does not by itself
establish real-world clinical benefit, which remains a limitation of in-silico
validation (Sec.~\ref{sec:limitations}); the retrospective case studies in
Appendix~\ref{app:case_study} provide complementary, predictor-independent
evidence of clinical plausibility.

\subsection{Full Ablation Study}
\label{sec:full_ablation}
This section provides the task-specific ablation results complementing the pooled analysis in \Cref{tab:agent_ablation}. As shown in \Cref{tab:agent_ablation_by_task}, the full system consistently achieves the highest Iter5 improvement across all three failure modes. Each ablated variant leads to a statistically significant drop, indicating that memory, pool-based exploitation, and structured analysis each contribute to the redesign performance across different trial failure types.

\begin{table*}[h]
\centering
\caption{Ablation study of the \mname\ framework on three failure modes: enrollment, safety, and efficacy. The primary metric is Iter5 cumulative improvement. For each ablated variant, we report the one-sided paired $t$-test against the full system with alternative hypothesis $H_1:\Delta p_5^{\mathrm{ablation}} < \Delta p_5^{\mathrm{full}}$. Effect sizes are reported as Cohen's $d_z$. Significance levels are denoted by *$p<0.05$, **$p<0.01$, and ***$p<0.001$.}
\label{tab:agent_ablation_by_task}
\begin{adjustbox}{width=0.7\linewidth,center}
\begin{tabular}{@{}llcccc@{}}
\toprule
\textbf{Task} & \textbf{Method} & \textbf{N} & \textbf{Iter5 $\Delta p$} & \textbf{$p$-value} & \textbf{Cohen's $d_z$} \\
\midrule

\multirow{4}{*}{Enrollment}
& Full System  & 20 & $+0.1032 {\scriptstyle \pm 0.0438}$ & --- & --- \\
& w/o Memory   & 20 & $+0.0624 {\scriptstyle \pm 0.0454}$ & $(2.289{\times}10^{-6})^{***}$ & $1.41$ \\
& w/o Pool     & 20 & $+0.0503 {\scriptstyle \pm 0.0392}$ & $(7.035{\times}10^{-8})^{***}$ & $1.81$ \\
& w/o Analysis & 20 & $+0.0540 {\scriptstyle \pm 0.0338}$ & $(1.131{\times}10^{-8})^{***}$ & $2.04$ \\

\midrule

\multirow{4}{*}{Safety}
& Full System  & 20 & $+0.0613 {\scriptstyle \pm 0.0436}$ & --- & --- \\
& w/o Memory   & 20 & $+0.0521 {\scriptstyle \pm 0.0464}$ & $0.002^{**}$ & $0.72$ \\
& w/o Pool     & 20 & $+0.0500 {\scriptstyle \pm 0.0443}$ & $(7.632{\times}10^{-5})^{***}$ & $1.05$ \\
& w/o Analysis & 20 & $+0.0545 {\scriptstyle \pm 0.0446}$ & $0.009^{**}$ & $0.58$ \\

\midrule

\multirow{4}{*}{Efficacy}
& Full System  & 20 & $+0.0578 {\scriptstyle \pm 0.0457}$ & --- & --- \\
& w/o Memory   & 20 & $+0.0450 {\scriptstyle \pm 0.0394}$ & $0.011^{*}$ & $0.56$ \\
& w/o Pool     & 20 & $+0.0458 {\scriptstyle \pm 0.0415}$ & $0.013^{*}$ & $0.54$ \\
& w/o Analysis & 20 & $+0.0480 {\scriptstyle \pm 0.0412}$ & $0.029^{*}$ & $0.45$ \\

\bottomrule
\end{tabular}
\end{adjustbox}
\end{table*}

\section{Deferred Full Case Study}\label{app:case_study}
\label{sec:case_study_detail}

We further validate \mname\ Agent's reasoning against additional real-world protocol redesigns, including cases where the agent correctly diagnosed an issue but could not express it as a typed augmentation.

\paragraph{Poor Enrollment, Mapracorat (NCT01298752 $\rightarrow$ NCT01591161)} 

We analyze NCT01298752, a Phase 3 trial of Mapracorat (anti-inflammatory ophthalmic suspension) for post-cataract surgery inflammation that failed due to poor enrollment. Sponsored by Bausch \& Lomb, the trial was subsequently redesigned and successfully executed as NCT01591161. Table~\ref{tab:case_study_train_mapracorat} compares the real-world redesign with \mname\ Agent's proposals.

\begin{table*}[h]
\centering
\caption{Real-world validation (NCT01298752, Poor Enrollment): We compare the real-world Mapracorat redesign with \mname\ Agent's proposed modifications, categorizing alignment as: \checkmark (match), $\sim$ (strategic alignment, tactical differences), or $\times$ (missed or incorrect). Rewards are the simulation-environment reward of the top-ranked augmentation supporting each row.}
\label{tab:case_study_train_mapracorat}
\small
\resizebox{\textwidth}{!}{%
\begin{tabular}{@{}p{2.8cm}p{4.7cm}p{5.1cm}cp{2.6cm}@{}}
\toprule
\textbf{Change Type} & \textbf{Real-World Redesign} & \textbf{\mname\ Agent Proposal} & \textbf{Match} & \textbf{Impact Level} \\
\midrule
\multicolumn{5}{l}{\textit{\textbf{Major Redesigns (Critical to Enrollment Success)}}} \\
\midrule
Enrollment Window & SHIFTED enrollment from \textbf{pre-operative} to \textbf{postoperative Day 1}, screening patients only after surgery had been performed & ADDED an inclusion criterion requiring \textbf{cataract surgery within the last 48 hours}, with postoperative inflammation and pain as the eligibility signal \textbf{reward: +0.023} & $\sim$ & Major, primary fix that dissolves the standard-of-care conflict \\
\midrule
Standard-of-Care Conflict & REMOVED the exclusion of \textbf{systemic or ocular corticosteroids within 14 days prior to surgery}, allowing investigators to keep their routine pre-operative protocol & DIAGNOSED the pre-procedure medication restriction and the workflow timing mismatch as the enrollment barrier, but produced \textbf{no exclusion-criterion modification} to remove it & $\sim$ & Major, the mechanism that deterred investigators \\
\midrule
Inflammation Enrichment & ADDED \textbf{$\geq$ Grade 2 (6--15 cells) anterior chamber cells} in the study eye at postoperative day 1 & PROPOSED baseline severity enrichment at \textbf{Grade 1} AC cells and pain within 48 hours, and in another augmentation, AC cell Grade 1–2 within 14 days, \textbf{reward: +0.002} & $\sim$ & Major, assay sensitivity for the inflammation endpoint \\
\midrule
Fellow-Eye Scheduling & REMOVED the inclusion clause requiring subjects to \textbf{delay fellow-eye cataract surgery} until study completion & DELETED the same clause in three separate optimization cycles and additionally proposed explicitly permitting fellow-eye surgery at any time \textbf{reward: +0.007} & \checkmark & Moderate, removes a scheduling deterrent \\
\midrule
Dosing Frequency & INCREASED dosing from \textbf{once daily (QD)} to \textbf{four times daily (QID)} over the same 14-day treatment period & Missing: no dosage augmentation was proposed at any optimization cycle & $\times$ & Moderate, raises ocular exposure supporting the efficacy signal \\
\midrule
\multicolumn{5}{l}{\textit{\textbf{Deliberate Preservations}}} \\
\midrule
Study Drug and Duration & PRESERVED mapracorat ophthalmic suspension 3\% and the 14-day treatment period & Preserved; all proposed modifications were confined to inclusion criteria & \checkmark & Critical, the redesign must test the same agent \\
\midrule
Primary Endpoints & PRESERVED both endpoints: percentage of participants with complete AC cell resolution and with Grade 0 pain, at day 8 & Preserved; no endpoint modification was proposed & \checkmark & Critical, keeps the two trials comparable \\
\midrule
Core Safety Exclusions & PRESERVED exclusions for severe or serious ocular conditions, chronic systemic disease, and study-drug hypersensitivity & Preserved; no exclusion criterion was relaxed or removed & \checkmark & Important, protects vulnerable surgical patients \\
\bottomrule
\end{tabular}%
}
\end{table*}

The enrollment barrier in the failed protocol was not a narrow patient definition but a conflict with surgical routine: enrolling before surgery meant enforcing a 14-day pre-operative corticosteroid washout, and few surgeons were willing to withhold standard care to recruit. The expert redesign resolved this structurally rather than by relaxing thresholds, moving the entire enrollment decision to postoperative day 1, at which point pre-operative medication is no longer a constraint and confirmed $\geq$ Grade 2 anterior chamber cells can serve as an enrichment criterion. \mname\ Agent reproduced the structural move: its highest-reward proposal restricts eligibility to patients already operated on within 48 hours and makes postoperative inflammation the screening signal, and it independently removed the fellow-eye scheduling clause across three optimization cycles. Its reasoning chain also named the pre-procedure medication restriction as the barrier. The agent set the enrichment threshold at Grade 1 rather than Grade 2 and its 48-hour window is looser than the expert's day-1 assessment, so the enriched population would be more heterogeneous than intended; it never converted its own diagnosis of the corticosteroid conflict into a modification of the exclusion criteria, because its accepted actions were confined to inclusion criteria; and it did not propose the increase from once-daily to four-times-daily dosing, nor the two operational refinements the expert added, a postoperative visual acuity feasibility screen and standardized baseline inflammation limits at the screening visit.

\paragraph{Poor Enrollment, Mircera (NCT00576602 $\rightarrow$ NCT01342640)}

We analyze NCT00576602, a Phase 4 trial of Mircera (methoxy polyethylene glycol-epoetin beta) for anemia in chronic kidney disease after kidney transplantation, sponsored by Hoffmann-La Roche, which terminated for poor recruitment. The failed protocol randomized a narrow post-transplant population. The successful redesign (NCT01342640) moved the same drug into the broader pre-dialysis CKD population under a single-arm open-label design. Table~\ref{tab:case_study_train_mircera} compares the real-world redesign with \mname\ Agent's proposals.

\begin{table*}[h]
\centering
\caption{Real-world validation (NCT00576602, Poor Enrollment): We compare the real-world Mircera redesign with \mname\ Agent's proposed modifications, categorizing alignment as: \checkmark (match), $\sim$ (strategic alignment, tactical differences), or $\times$ (missed or incorrect). Rewards are the simulation-environment reward of the top-ranked augmentation supporting each row.}
\label{tab:case_study_train_mircera}
\small
\resizebox{\textwidth}{!}{%
\begin{tabular}{@{}p{2.8cm}p{4.7cm}p{5.1cm}cp{2.6cm}@{}}
\toprule
\textbf{Change Type} & \textbf{Real-World Redesign} & \textbf{\mname\ Agent Proposal} & \textbf{Match} & \textbf{Impact Level} \\
\midrule
\multicolumn{5}{l}{\textit{\textbf{Major Redesigns (Critical to Enrollment Success)}}} \\
\midrule
Population Scope & PIVOTED from \textbf{post-transplant CKD} patients 6 months to 5 years after transplantation to the far larger \textbf{pre-dialysis CKD} population routinely seen in nephrology clinics & DIAGNOSED the narrow post-transplant population as the enrollment barrier, but every proposed action remained inside it; the widest was extending the transplant window to \textbf{6 months to 10 years} \textbf{reward: +0.029} & $\sim$ & Major, primary fix that moves recruitment into an accessible clinic population \\
\midrule
Renal Function Definition & ADDED the core eligibility axis of the new population: \textbf{not on dialysis} with \textbf{eGFR 20--60 mL/min} & Missing: no renal function criterion was proposed, as the failed protocol contained no renal function row for the agent to modify & $\times$ & Major, operationally defines the accessible patient pool \\
\midrule
Comparator Structure & REPLACED the two-arm randomization against \textbf{no ESA therapy} with a \textbf{single-arm, open-label} design in which every patient receives Mircera & IDENTIFIED the no-treatment control arm as an ethical and recruitment barrier for anemic patients, but produced \textbf{no design-level modification}; all accepted actions were eligibility edits & $\sim$ & Major, removes the ethical objection that deterred both patients and investigators \\
\midrule
Dosing Regimen & CHANGED dosing from \textbf{0.6 mcg/kg every 2 weeks} to \textbf{1.2 mcg/kg every 4 weeks}, doubling the dose and moving to monthly administration & Missing: no dosage augmentation was proposed at any optimization cycle & $\times$ & Moderate, showcases the once-monthly schedule that distinguishes the drug \\
\midrule
Anemia Severity Threshold & REQUIRED \textbf{Hb $<$ 10 g/dL} at baseline, replacing the undefined ``anemia'' criterion & PRODUCED the same threshold, requiring \textbf{baseline Hb below 10 g/dL}, \textbf{reward: +0.006}; its higher-reward anemia definitions move the opposite way, to Hb $<$ 12--13 g/dL and to a 10.0--13.5 g/dL band that would exclude the target population & $\sim$ & Moderate, enriches for patients able to demonstrate an ESA response \\
\midrule
\multicolumn{5}{l}{\textit{\textbf{Deliberate Preservations}}} \\
\midrule
ESA Washout & PRESERVED the requirement of \textbf{no ESA therapy during the 3 months} before study start, so that the hemoglobin response is attributable to Mircera & Proposed \textbf{relaxing} the washout in two augmentations, admitting recently ESA-treated patients on a hemoglobin-stability condition \textbf{reward: +0.010} & $\times$ & Critical, recent ESA exposure would confound the primary hemoglobin endpoint \\
\bottomrule
\end{tabular}%
}
\end{table*}

The failed protocol's criteria were written too tightly; it failed because the population it defined is intrinsically hard to reach. Post-transplant patients are managed by transplant teams reluctant to add an investigational agent, and asking anemic patients to accept a no-treatment arm compounded the problem. The expert redesign therefore abandoned the population rather than relaxing it, moving the same drug into pre-dialysis CKD, where patients are routinely seen in nephrology clinics, and converting the study to a single-arm open-label design so that no participant was randomized away from treatment. \mname\ Agent reached both halves of this diagnosis: its reasoning chain names the narrow post-transplant population and the ethical barrier of the no-ESA control arm, and it correctly generated the expert's Hb $<$ 10 g/dL severity threshold. What it could not do was act on either insight. Its accepted actions were confined to editing eligibility criteria of the existing protocol, so the population pivot became a widening of the transplant window to ten years, the comparator diagnosis produced no design change at all, and the renal function axis that defines the successful population was never proposed because the failed protocol contained no such criterion to modify. The same confinement explains further misses: the doubled monthly dose and the shift of the primary hemoglobin assessment from Weeks 13--16 to Weeks 20, 24 and 28. Those misses are silences, axes on which the agent proposed nothing at all. The one axis on which it moved actively against the expert is instructive rather than harmful: relaxing the three-month ESA washout would broaden recruitment, but recent ESA exposure would confound exactly the hemoglobin response the trial is designed to measure, and the expert retained the washout for that reason.

\paragraph{Poor Enrollment, Stannsoporfin (NCT00850993 $\rightarrow$ NCT01887327)}

We analyze NCT00850993, a Phase 2 trial of stannsoporfin, a heme oxygenase inhibitor, for neonatal hyperbilirubinemia, sponsored by InfaCare Pharmaceutical, which terminated with the registered reason ``to redefine study population''. The failed protocol enrolled newborns merely at risk of hemolytic disease and required serum bilirubin below the phototherapy threshold, recruiting low-risk infants for whom an investigational drug was difficult to justify to parents. The successful redesign (NCT01887327) enrolled only newborns with confirmed hemolysis indicators who already required phototherapy, and gave phototherapy to every arm. Table~\ref{tab:case_study_train_stannsoporfin} compares the real-world redesign with \mname\ Agent's proposals.

\begin{table*}[h]
\centering
\caption{Real-world validation (NCT00850993, Poor Enrollment): We compare the real-world stannsoporfin redesign with \mname\ Agent's proposed modifications, categorizing alignment as: \checkmark (match), $\sim$ (strategic alignment, tactical differences), or $\times$ (missed or incorrect). Rewards are the simulation-environment reward of the top-ranked augmentation supporting each row.}
\label{tab:case_study_train_stannsoporfin}
\small
\resizebox{\textwidth}{!}{%
\begin{tabular}{@{}p{2.8cm}p{4.7cm}p{5.1cm}cp{2.6cm}@{}}
\toprule
\textbf{Change Type} & \textbf{Real-World Redesign} & \textbf{\mname\ Agent Proposal} & \textbf{Match} & \textbf{Impact Level} \\
\midrule
\multicolumn{5}{l}{\textit{\textbf{Major Redesigns (Critical to Enrollment Success)}}} \\
\midrule
Hemolysis Enrichment & REPLACED the unoperationalized ``at risk for hemolytic disease'' criterion with \textbf{ABO or Rh incompatibility confirmed by a positive Coombs test, or G6PD deficiency}, and inverted the bilirubin requirement from \textbf{below} to \textbf{at or above} the age-specific phototherapy threshold & PROPOSED requiring \textbf{Rh or ABO incompatibility with baseline TSB $\geq$ 9 mg/dL}, and separately a confirmed hemolytic disease diagnosis with bilirubin $\geq$ 6 mg/dL; both enrich rather than broaden, but neither is anchored to the phototherapy guideline \textbf{reward: +0.005} & $\sim$ & Major, primary fix that concentrates clinical need and pharmacological effect \\
\midrule
Phototherapy Backbone & MADE standard phototherapy \textbf{mandatory for every arm}, started within 30 minutes of injection, so that randomization never withholds standard care & DIAGNOSED that families prefer established phototherapy and that a placebo-controlled design in newborns raises ethical concern, but \textbf{no augmentation required phototherapy across arms} & $\sim$ & Major, removes the ethical objection that deterred parental consent \\
\midrule
\multicolumn{5}{l}{\textit{\textbf{Deliberate Preservations}}} \\
\midrule
Neonatal Maturity Floor & PRESERVED birth weight \textbf{$\geq$ 2500 g} and term or near-term gestational age (35--43 weeks) & Proposed \textbf{lowering birth weight to 1800 g} \textbf{reward: +0.019} and, in three further augmentations, extending gestational age down to \textbf{28--34 weeks} & $\times$ & Critical, preterm and low-birth-weight neonates carry distinct bilirubin and safety risks \\
\midrule
Cardiac Safety Exclusions & PRESERVED the exclusions for \textbf{QT-prolonging medications} and for long-QT and porphyria family history, and expanded the safety envelope overall from 4 to 25 exclusion criteria & Proposed \textbf{admitting neonates on QT-prolonging medications} under prescriber confirmation --- the \textbf{highest-reward action of all eleven} \textbf{reward: +0.021} --- and separately admitting infants with long-QT or porphyria family history & $\times$ & Critical, stannsoporfin is photoreactive and the population cannot report cardiac symptoms \\
\midrule
Major Abnormality Exclusion & PRESERVED the exclusion of neonates or mothers with abnormalities or infections that could compromise safety or confound the analysis & Preserved: the agent edited three of the four exclusion criteria but left this one intact & \checkmark & Important, the broadest safety guardrail in a neonatal protocol \\
\bottomrule
\end{tabular}%
}
\end{table*}

The failed protocol recruited the wrong newborns for the wrong reason. By requiring bilirubin below the phototherapy threshold it selected infants who did not yet need treatment, which made an investigational injection hard to justify to parents and diluted the pharmacological signal the drug was designed to produce. The expert redesign inverted both halves of that logic: it enrolled only newborns with laboratory-confirmed hemolysis, ABO or Rh incompatibility with a positive Coombs test or G6PD deficiency, who had already crossed the phototherapy threshold, and it guaranteed phototherapy to every participant so that no family was asked to accept the possibility of no treatment at all. \mname\ Agent recovered the enrichment logic. Two of its augmentations require blood-type incompatibility or a confirmed hemolytic diagnosis together with a minimum baseline bilirubin, which is the correct direction and a reversal of the broadening its other augmentations favor, and its reasoning chain names both the ethical concern of a placebo-controlled neonatal design and the parental preference for established care. Two limits are visible. First, the correct proposals were not promoted: they rank seventh and eleventh of eleven, while the reward model favored changes that widen the population. Second, and more consequentially, that widening ran against the expert's safety judgment. The agent's highest-reward action of all admits neonates receiving QT-prolonging medications, and it further proposed lowering the birth-weight floor to 1800 grams, extending gestational age into preterm ranges, and dropping the long-QT and porphyria family-history exclusion.

\paragraph{Poor Enrollment, Daprodustat (NCT01406340 $\rightarrow$ NCT02243306)}

We analyze NCT01406340, a Phase 1 pharmacokinetic study of daprodustat (GSK1278863) in renal impairment, sponsored by GlaxoSmithKline, terminated for ``difficulty enrolling last cohort of subjects''. The failed protocol required parallel recruitment into four renal-function cohorts spanning normal function through Stage 5 hemodialysis, each carrying its own eligibility block, so that a single stalled stratum could halt the entire study. The successful redesign (NCT02243306) reduced the study to one standalone cohort of end-stage renal disease patients on peritoneal dialysis. Table~\ref{tab:case_study_train_daprodustat} compares the real-world redesign with \mname\ Agent's proposals.

\begin{table*}[h]
\centering
\caption{Real-world validation (NCT01406340, Poor Enrollment): We compare the real-world daprodustat renal-PK redesign with \mname\ Agent's proposed modifications, categorizing alignment as: \checkmark (match), $\sim$ (strategic alignment, tactical differences), or $\times$ (missed or incorrect). Rewards are the simulation-environment reward of the top-ranked augmentation supporting each row.}
\label{tab:case_study_train_daprodustat}
\small
\resizebox{\textwidth}{!}{%
\begin{tabular}{@{}p{2.8cm}p{4.7cm}p{5.1cm}cp{2.6cm}@{}}
\toprule
\textbf{Change Type} & \textbf{Real-World Redesign} & \textbf{\mname\ Agent Proposal} & \textbf{Match} & \textbf{Impact Level} \\
\midrule
\multicolumn{5}{l}{\textit{\textbf{Major Redesigns (Critical to Enrollment Success)}}} \\
\midrule
Cohort Structure & DECOUPLED the \textbf{four parallel renal-function cohorts} (normal, Stage 3, Stage 4, Stage 5 hemodialysis) into a \textbf{single standalone cohort}, removing the dependency by which one stalled stratum terminated the study & DIAGNOSED the failure precisely as multi-subgroup structural fragility and the logistical burden of simultaneous recruitment across distinct populations, but proposed \textbf{no cohort-level action}; its only renal-function move relaxed a creatinine clearance threshold to $\geq$ 30 mL/min \textbf{inside} the existing four-cohort design \textbf{reward: +0.004} & $\sim$ & Major, primary fix that removes the single point of failure \\
\midrule
Dialysis Modality & SWITCHED the target population from stable \textbf{in-centre hemodialysis} three times weekly to \textbf{peritoneal dialysis for at least 2 months}, an ambulatory home-based group far more available for outpatient pharmacokinetic sampling & Missing: no dialysis modality change was proposed at any optimization cycle & $\times$ & Major, selects the population that can realistically be recruited \\
\midrule
Criteria Structure & REPLACED the per-cohort eligibility blocks with a \textbf{flat safety and efficacy structure} for one population, eliminating the cohort-matching burden & INVERTED the intended target: \textbf{seven of thirteen augmentations edited section headers and preamble rows} such as ``all study participants'' and ``additional exclusion criteria for subjects with normal renal function'' rather than clinical criteria, five of them earning \textbf{reward: +0.001} or less & $\times$ & Moderate, removes protocol complexity that slowed site activation \\
\midrule
\multicolumn{5}{l}{\textit{\textbf{Deliberate Preservations}}} \\
\midrule
Screening Laboratory Floor & PRESERVED the requirement that \textbf{vitamin B12 and folate} be above the lower limit of normal, and tightened laboratory screening further with AST, ALT and bilirubin $\leq$ 1.5 $\times$ ULN & Proposed \textbf{relaxing} the vitamin B12 and folate floor in two augmentations, and separately widened the hemoglobin window and raised the urinalysis blood allowance; the hemoglobin relaxation was its \textbf{highest-reward action} \textbf{reward: +0.013} & $\times$ & Important, nutritional anemia would confound an erythropoiesis-targeted pharmacokinetic study \\
\bottomrule
\end{tabular}%
}
\end{table*}

This case separates diagnosis from action more sharply than any other in the appendix. \mname\ Agent read the failure correctly and completely: its reasoning chain identifies multi-subgroup structural fragility and the logistical burden of recruiting four matched renal-function cohorts in parallel, which is exactly why the study died when its last cohort would not fill. The expert acted on that same diagnosis by deleting the structure, running one standalone cohort of end-stage renal disease patients and moving from in-centre hemodialysis to peritoneal dialysis, a population that is ambulatory, home-based, and therefore practical to sample repeatedly in a Phase 1 protocol. The agent could not take that step. Its accepted actions were confined to editing individual eligibility criteria of the existing protocol, and a cohort is not an eligibility criterion, so a correct structural diagnosis had nowhere to go. What emerged instead was a set of peripheral relaxations, a wider hemoglobin window, a higher urinalysis blood allowance, and a lower vitamin B12 and folate floor, the last of which works against the study, since nutritional anemia is precisely the confounder a hypoxia-inducible factor prolyl hydroxylase inhibitor study needs to exclude. Two further symptoms indicate that the action layer, not the reasoning layer, is the bottleneck. Seven of the thirteen proposals modify section headers and preamble text rather than clinical content, and one proposal adds a Mini-Mental State Examination threshold, introducing a new screening barrier into a trial that had already failed to enroll.

\paragraph{Efficacy Inadequacy, Brivaracetam (NCT00699283 $\rightarrow$ NCT01261325)}

We analyze NCT00699283, a Phase 3 conversion-to-monotherapy study of brivaracetam in adults with uncontrolled partial-onset seizures, sponsored by UCB Pharma, stopped at interim analysis for lack of efficacy with 62 subjects enrolled. The failed protocol tapered off each patient's background antiepileptic drugs so that brivaracetam stood alone, and measured the cumulative exit rate at 112 days against a fixed historical-control threshold of 0.722. The successful redesign (NCT01261325) kept background therapy in place and tested brivaracetam as add-on against a concurrent placebo. Table~\ref{tab:case_study_train_brivaracetam} compares the real-world redesign with \mname\ Agent's proposals.

\begin{table*}[h]
\centering
\caption{Real-world validation (NCT00699283, Efficacy Inadequacy): We compare the real-world brivaracetam redesign with \mname\ Agent's proposed modifications, categorizing alignment as: \checkmark (match), $\sim$ (strategic alignment, tactical differences), or $\times$ (missed or incorrect). Rewards are the simulation-environment reward of the top-ranked augmentation supporting each row.}
\label{tab:case_study_train_brivaracetam}
\small
\resizebox{\textwidth}{!}{%
\begin{tabular}{@{}p{2.8cm}p{4.7cm}p{5.1cm}cp{2.6cm}@{}}
\toprule
\textbf{Change Type} & \textbf{Real-World Redesign} & \textbf{\mname\ Agent Proposal} & \textbf{Match} & \textbf{Impact Level} \\
\midrule
\multicolumn{5}{l}{\textit{\textbf{Major Redesigns (Critical to Efficacy Success)}}} \\
\midrule
Treatment Paradigm & SHIFTED from conversion-to-monotherapy with an active background AED taper to \textbf{adjunctive brivaracetam on 1--2 concomitant AEDs kept stable at optimal dose} throughout baseline and treatment & Missing: the monotherapy taper was never questioned; all seven augmentations modify dose or eligibility while treating the standalone regimen as fixed & $\times$ & Major, the linchpin change: background therapy is both a retention safety net during seizure breakthrough and the condition for an interpretable add-on efficacy signal \\
\midrule
Comparator and Endpoint & REPLACED the \textbf{fixed historical-control threshold of 0.722} on cumulative exit rate at 112 days with a \textbf{concurrent placebo comparison} on percent reduction in seizure frequency and 50\% responder rate over 12 weeks & DIAGNOSED the external benchmark as too ambitious and proposed a \textbf{concurrent control with an event-frequency reduction endpoint} & $\sim$ & Major, changes the statistical paradigm from clearing an external bar to beating a concurrent arm \\
\midrule
Dose Range & MOVED from 50 and 100 mg/day to \textbf{100 and 200 mg/day}, dropping the lowest arm and adding a higher-exposure arm & DROPPED the 50 mg arm in all three dosage augmentations and moved upward to \textbf{100, 150 and 200 mg/day} \textbf{reward: +0.072} & \checkmark & Major, higher total exposure improves the chance of a dose-responsive signal \\
\midrule
Baseline Enrichment & ADDED a seizure-frequency floor of \textbf{$\geq$ 8 Type I seizures during the 8-week baseline}, with at least 2 in each 4-week interval & PROPOSED \textbf{$\geq$ 4 focal seizures per month} over three months, the same magnitude as the expert criterion, but the reward model ranked it \textbf{last of seven} \textbf{reward: +0.001} & $\sim$ & Moderate, guarantees enough baseline events for a treatment effect to be measurable \\

\bottomrule
\end{tabular}%
}
\end{table*}

The failed trial set itself an externally anchored target and then removed the support its patients needed to reach it. Tapering background antiepileptic drugs left brivaracetam to prevent seizures alone while the cumulative exit rate was compared against a fixed benchmark derived from other agents, so patients who broke through during the taper counted directly against the drug. The expert redesign changed the regimen and the comparator together, keeping concomitant AEDs stable and testing brivaracetam as add-on against a concurrent placebo on seizure-frequency reduction. \mname\ Agent got the pharmacology right and the architecture wrong. All three of its dosage proposals dropped the 50 mg arm and moved upward to between 100 and 200 mg per day, matching the direction the sponsor took, and its analysis independently identified the historical-control benchmark as an unrealistic target and called for a concurrent comparator with an event-frequency endpoint. But those endpoint proposals did not survive validation, and the regimen itself was never treated as modifiable, so the single change that made the successful trial possible, keeping the patient's existing therapy in place, was never considered. The eligibility proposals show the same pattern in miniature. The agent did produce the expert's enrichment criterion, requiring at least four focal seizures per month, but the reward model scored it near zero and ranked it last.

\paragraph{Efficacy Inadequacy, Intranasal Dexmedetomidine (NCT02169336 $\rightarrow$ NCT02284243)}

We analyze NCT02169336, a Phase 2 trial of intranasal dexmedetomidine (DEX-IN) for acute postoperative pain following bunionectomy, sponsored by Baudax Bio, terminated at interim analysis for lack of efficacy. The failed protocol spread 95 participants across three arms, DEX-IN 35 mcg, DEX-IN 50 mcg and placebo, and extended dosing beyond the fixed 48-hour efficacy window with an as-needed phase of up to three further days. The successful redesign (NCT02284243) kept the drug, the endpoint and the patients unchanged and altered only the trial's arithmetic. Table~\ref{tab:case_study_train_efficacy_dex} compares the real-world redesign with \mname\ Agent's proposals.

\begin{table*}[h]
\centering
\caption{Real-world validation (NCT02169336, Efficacy Inadequacy): We compare the real-world dexmedetomidine redesign with \mname\ Agent's proposed modifications, categorizing alignment as: \checkmark (match), $\sim$ (strategic alignment, tactical differences), or $\times$ (missed or incorrect). Rewards are the simulation-environment reward of the top-ranked augmentation supporting each row.}
\label{tab:case_study_train_efficacy_dex}
\small
\resizebox{\textwidth}{!}{%
\begin{tabular}{@{}p{2.8cm}p{4.7cm}p{5.1cm}cp{2.6cm}@{}}
\toprule
\textbf{Change Type} & \textbf{Real-World Redesign} & \textbf{\mname\ Agent Proposal} & \textbf{Match} & \textbf{Impact Level} \\
\midrule
\multicolumn{5}{l}{\textit{\textbf{Major Redesigns (Critical to Efficacy Success)}}} \\
\midrule
Statistical Power & INCREASED enrolment from \textbf{95 to 168 participants} across two rather than three arms, raising per-arm sample size roughly \textbf{2.65-fold} & DIAGNOSED the root cause as \textbf{underpowering} and named a target of 100 participants in its analysis, but emitted \textbf{no augmentation}: sample size has no corresponding action & $\sim$ & Major, primary fix; the trial was unable to detect analgesia even if the drug worked \\
\midrule
Arm Consolidation & DROPPED the \textbf{35 mcg arm} entirely, concentrating power on the single dose that had previously shown the strongest signal & DROPPED the 35 mcg arm in three of four dosage proposals, but each substituted a different regimen change --- a 60 mcg loading dose, q4h instead of q6h, or continuous intranasal infusion --- rather than a clean two-arm consolidation, and a fourth retained 35 mcg \textbf{reward: +0.038} & $\sim$ & Major, converts wasted allocation into statistical power \\
\midrule
PRN Elimination & REMOVED the \textbf{as-needed dosing phase of up to 3 additional days}, confining treatment to the fixed 48-hour window the endpoint measures & Missing: \textbf{every} dosage proposal retained the PRN extension, and one lengthened it to \textbf{5 days} & $\times$ & Moderate, PRN dosing dilutes the fixed 48-hour efficacy signal it sits outside \\
\midrule
\multicolumn{5}{l}{\textit{\textbf{Deliberate Preservations}}} \\
\midrule
Primary Endpoint & PRESERVED \textbf{SPID48} with identical Numeric Rating Scale time points, keeping the two trials directly comparable & CONTRADICTED itself: the structured diagnosis states that \textbf{SPID48 is appropriate}, yet the \textbf{highest-reward action replaces it} --- with SPID24, with an SPID24 and PID6 composite, or with pain-satisfaction composites \textbf{reward: +0.055} & $\times$ & Important, SPID48 is the standard acute-pain endpoint and the axis on which the redesign is comparable \\
\midrule
Study Drug and Route & PRESERVED \textbf{intranasal dexmedetomidine} at 50 mcg as the investigational agent & Preserved: the agent rewrote the dosage aspect four times and never changed the molecule or the intranasal route & \checkmark & Critical, the redesign exists to give the same drug a fair test \\
\bottomrule
\end{tabular}%
}
\end{table*}

This redesign is unusually instructive because of how little it changed. The sponsor kept the drug, the endpoint, the eligibility criteria, the masking, and fixed the trial purely by arithmetic: one fewer arm, no as-needed phase, and enough patients per arm to see an effect. \mname\ Agent identified the failure correctly and reproducibly, returning underpowering as the primary root cause and naming a required sample size in its analysis, and it reached part of the fix, dropping the 35 mcg arm in three of its four dosing proposals. Beyond that its behaviour diverges from the expert in two distinct ways. The first is representational: the trial's central fix, enrolling more patients, is not expressible in the agent's action vocabulary, so a correct recommendation stayed in prose and never became a proposal. The second is more interesting, because the agent's own analysis was right and its action layer overrode it. The structured diagnosis states that SPID48 is an appropriate endpoint, yet the highest-reward action in every run rewrites it. The agent also retained the as-needed dosing phase in every planning phase and in one case extended it, leaving in place the design element that diluted the very window the endpoint measures. Where the expert succeeded by subtracting, the agent consistently added.

\paragraph{Safety/Adverse Events, EXPAREL (NCT01919190 $\rightarrow$ NCT02199574)}

We analyze NCT01919190, a Phase 4 trial of EXPAREL (liposomal bupivacaine) via transversus abdominis plane (TAP) infiltration for post-surgical pain in lower abdominal procedures that failed due to drug adverse events. Sponsored by Pacira Pharmaceuticals, the drug was subsequently redesigned and successfully executed as NCT02199574 in a different surgical context. Table~\ref{tab:case_study_train_safety_exparel} compares the real-world redesign with \mname\ Agent's proposals.

\begin{table*}[h]
\centering
\caption{Real-world validation (NCT01919190, Safety/Adverse Events): We compare the real-world EXPAREL redesign with \mname\ Agent's proposed modifications, categorizing alignment as: \checkmark (match), $\sim$ (strategic alignment, tactical differences), or $\times$ (missed or incorrect).}
\label{tab:case_study_train_safety_exparel}
\small
\resizebox{\textwidth}{!}{%
\begin{tabular}{@{}p{2.8cm}p{4.7cm}p{5.1cm}cp{2.6cm}@{}}
\toprule
\textbf{Change Type} & \textbf{Real-World Redesign} & \textbf{\mname\ Agent Proposal} & \textbf{Match} & \textbf{Impact Level} \\
\midrule
\multicolumn{5}{l}{\textit{\textbf{Major Redesigns (Critical to Safety Success)}}} \\
\midrule
Trial Type \& Primary Outcome & PIVOTED to \textbf{PK/safety} characterization: original failed trial attempted to prove opioid-sparing efficacy and improved OBAS scores in a heterogeneous surgical population & MODIFIED to \textbf{PK/safety}: ``Evaluate plasma levels of bupivacaine and safety metrics following a single administration of EXPAREL'' & \checkmark & Fundamental redesign addressing the root cause \\
\midrule
Dosage Reduction & \textbf{REDUCED by 50\%}: 266 mg/20 mL (60 mL total volume) $\rightarrow$ 133 mg/10 mL single dose & \textbf{REDUCED by $\sim$50\%}: proposed 133 mg in 20 mL saline per validated option (40 mL total volume) & \checkmark & Major, halves systemic bupivacaine exposure \\
\midrule
Surgical Model & CHANGED procedure entirely: lower abdominal surgeries with TAP infiltration $\rightarrow$ tonsillectomy with local surgical-site infiltration & Missing: agent did not propose leaving the hemorrhage-prone lower-abdominal/TAP context & $\times$ & Major, removes the anatomical context in which hemorrhage arose \\
\midrule
Eligibility Criteria & SIMPLIFIED: removed TAP-specific anatomical exclusions, complex surgical requirements, chronic opioid exclusions, pain-medication washout, metastatic disease, and substance-abuse exclusions; retained core safety exclusions & ADDED bleeding-specific exclusions and liver dysfunction criteria, while keeping the original complex exclusion structure & $\times$ & Minor, reduces protocol complexity around the new procedure \\
\midrule
\multicolumn{5}{l}{\textit{\textbf{Deliberate Preservations}}} \\
\midrule
Safety Focus & PRESERVED \textbf{key safety exclusions} such as local-anesthetic hypersensitivity, pregnancy/nursing, investigational-drug washout, and significant medical conditions & Correctly emphasized bleeding and systemic-exposure risk, but did so by \textbf{tightening exclusions} within the wrong surgical model & $\sim$ & Important, the exclusions that guard against local-anesthetic toxicity \\
\bottomrule
\end{tabular}%
}
\end{table*}

The primary safety issue in the failed trial was postoperative abdominal hemorrhage, plausibly linked to high-volume TAP infiltration in hemorrhage-prone lower-abdominal surgical sites. Both the real-world redesign and \mname\ Agent correctly identified two central safety moves: pivoting from an efficacy trial to PK/safety characterization and reducing the EXPAREL dose by approximately half. These are strong clinically plausible alignments because they directly address systemic bupivacaine exposure and safety monitoring. On the other hand, the expert redesign went further by changing the surgical model entirely, thereby removing the anatomical context in which hemorrhage had emerged. The agent instead tried to manage risk within the same broad surgical context by adding bleeding-specific restrictions. That proposal is clinically understandable, but it misses the structural safety move: eliminate the risky use context rather than merely screening around it.


\section{Deferred Full RAG-Grounded Case Studies}
\label{app:validator_cases}
\begin{tcolorbox}[title={Case 1: Disease-RAG prevents misaligned safety filtering in a Type 2 diabetes trial}, fontupper=\footnotesize, breakable]
\textbf{Clinical trial case.}
\emph{Trial:} NCT00294723. 
\emph{Condition:} Type 2 diabetes (liraglutide safety evaluation). 
\emph{Proposed redesign:} exclude patients with eGFR $< 60$ mL/min at screening.

\medskip
\textbf{Retrieved evidence (Disease-RAG).}
\emph{Query:} \texttt{Diabetes}. 
Retrieved guidelines emphasized management of \textbf{cardiovascular disease risk factors} as a core part of diabetes care.

\medskip
\textbf{Validation outcome.}
\emph{Decision:} \textbf{Invalid redesign.}

\medskip
\textbf{Reason.}
Combined with the adverse event summary, which included myocardial infarctions, the retrieved disease context showed that the proposed renal exclusion was overly broad and not well aligned with the trial's main safety risks. The Validator Agent therefore pruned the modification.
\end{tcolorbox}

\begin{tcolorbox}[title={Case 2: PubMed-RAG prevents unsafe diagnostic relaxation in a hypertension-in-pregnancy trial}, fontupper=\footnotesize, breakable]
\textbf{Clinical trial case.}
\emph{Trial:} NCT03872336. 
\emph{Condition:} Hypertension in pregnancy (labetalol treatment). 
\emph{Proposed redesign:} accept a single severe blood pressure reading ($\geq 160/110$ mmHg) for diagnosis.

\medskip
\textbf{Retrieved evidence (PubMed-RAG).}
\emph{Query:} \texttt{Hypertension diagnosis measurement criteria}. 
PMID 29574470 and PMID 24652215 both support diagnosis based on repeated blood pressure measurements over time.

\medskip
\textbf{Validation outcome.}
\emph{Decision:} \textbf{Invalid redesign.}

\medskip
\textbf{Reason.}
The Validator Agent determined that accepting a single severe reading is inconsistent with standard diagnostic practice and may cause false-positive diagnoses, unnecessary intervention, or inappropriate enrollment.
\end{tcolorbox}

\begin{tcolorbox}[title={Case 3: Drug-RAG preserves a critical safety exclusion in a labetalol trial}, fontupper=\footnotesize, breakable]
\textbf{Clinical trial case.}
\emph{Trial:} NCT03872336. 
\emph{Condition:} Hypertension in pregnancy (labetalol treatment). 
\emph{Proposed redesign context:} remove or weaken the exclusion criterion on persistent mild--moderate asthma.

\medskip
\textbf{Retrieved evidence (Drug-RAG).}
\emph{Query:} \texttt{labetalol toxicity}. 
Retrieved safety evidence noted risk of hypotension, bradycardia, and bronchospasm.

\medskip
\textbf{Validation outcome.}
\emph{Decision:} \textbf{Preserve existing criterion.}

\medskip
\textbf{Reason.}
The Validator Agent identified a drug--disease interaction between labetalol and asthma risk. Because bronchospasm is a foreseeable adverse effect, removing the asthma exclusion would be clinically unsafe.
\end{tcolorbox}

\section{Analyzer Agent Details}\label{app:analysis_agent}

The Analyzor Agent implements a domain-aware ReAct reasoning pipeline adapted by failure mode (enrollment, safety, efficacy). Novel components include adverse event profiling, statistical power assessment, and design-level pivots. Table~\ref{tab:analysis_variants} summarizes failure-mode-specific adaptations.

\begin{table}[h]
\centering
\caption{Analyzer Agent variants by failure mode.}
\label{tab:analysis_variants}
\small
\begin{tabular}{@{}p{3cm}p{3.5cm}p{3.5cm}p{3.5cm}@{}}
\toprule
\textbf{Component} & \textbf{Enrollment} & \textbf{Safety} & \textbf{Efficacy} \\
\midrule
\textbf{Profiling} & None & Adverse Event Profiling (severity, organ systems, root cause) & Efficacy Gap Profiling (observed vs expected) \\
\addlinespace
\textbf{Classification} & 4 categories & 5 categories (adds \texttt{safety\_inadequate}) & 4 categories (weights enrichment higher) \\
\addlinespace
\textbf{Assessments} & None & Dosage + AE profile & Dosage + Outcome + Power analysis \\
\addlinespace
\textbf{Prioritization} & Confidence-based & Safety-first (toxicity reduction priority) & Simplicity-first tiered (PRIMARY/SECONDARY/TERTIARY) \\
\bottomrule
\end{tabular}
\end{table}

\begin{tcolorbox}[title=Protocol Classification, colback=gray!5, fontupper=\footnotesize, breakable]
\textbf{Role:} Clinical researcher classifying eligibility criteria

\textbf{Task:} Classify criteria into 4 categories with confidence scores [0-1]

\textbf{Context:}
\begin{verbatim}
Phase: Phase 2
Mechanism: X inhibits Y pathway
Endpoint: Measuring Z at 12 weeks
\end{verbatim}

\textbf{Criteria to Classify:}
\begin{verbatim}
<criterion aspect_name="eligibility/inclusion_criteria" index="1">
Must wait for fellow eye surgery until study completion
</criterion>
<criterion aspect_name="eligibility/exclusion_criteria" index="2">
Any prior participation in drug trials within 12 months
</criterion>
\end{verbatim}

\textbf{Categories:}
\begin{enumerate}[noitemsep]
\item PARTICIPATION\_BARRIER: Timing/waiting requirements, administrative hurdles
\item SAFETY\_EXCLUSION: Medical risks (allergies, drug interactions, severe conditions)
\item SELECTION\_CRITERION: Defines WHO is eligible (disease type, procedure type, demographics)
\item ENRICHMENT\_CRITERION: Selects likely responders (biomarkers, mechanism-aligned traits)
\end{enumerate}

For each criterion, assign scores [0-1] to ALL categories, pick PRIMARY (highest), give 1-sentence reason.

\textbf{Output Format:}
\begin{verbatim}
<classification aspect_name="eligibility/inclusion_criteria" index="1">
<participation_barrier_score>0.92</participation_barrier_score>
<safety_exclusion_score>0.05</safety_exclusion_score>
<selection_criterion_score>0.20</selection_criterion_score>
<enrichment_criterion_score>0.10</enrichment_criterion_score>
<primary_category>PARTICIPATION_BARRIER</primary_category>
<reasoning>Waiting requirement for fellow eye surgery is a strong 
participation barrier with no medical justification.</reasoning>
</classification>
\end{verbatim}
\end{tcolorbox}

\begin{tcolorbox}[title=Mechanism Alignment Check, colback=gray!5, fontupper=\footnotesize, breakable]
\textbf{Role:} Clinical researcher evaluating mechanism alignment

\textbf{Task:} Check if criteria select mechanism-appropriate patients and detect missing enrichment

\textbf{Questions:}
\begin{enumerate}[noitemsep]
\item Do we select patients who HAVE the target condition this mechanism treats?
\item Do we select patients with baseline values allowing measurement of endpoint Y?
\item Are safety exclusions too broad, blocking potential responders?
\end{enumerate}

If missing enrichment (no criteria selecting treatment-responsive patients):
\begin{itemize}[noitemsep]
\item Propose ONE objective criterion with: measurement method, threshold, timing
\item Must be measurable (grades/scores/labs), not subjective ("anticipated"/"likely")
\end{itemize}

\textbf{Output Format:}
\begin{verbatim}
<mechanism_analysis>
Current criteria define cataract surgery candidates but lack enrichment 
for inflammation severity. Waiting requirement blocks eligible patients 
without medical benefit.
</mechanism_analysis>

<missing_enrichment_criterion>
Add inclusion: Baseline anterior chamber cell grade $\geq$2+ (SUN criteria) 
measured within 7 days of enrollment. Selects patients with measurable 
inflammation for mechanism-aligned response assessment.
</missing_enrichment_criterion>
\end{verbatim}
\end{tcolorbox}

\begin{tcolorbox}[title=Adverse Event Profiling, colback=gray!5, fontupper=\footnotesize, breakable]
\textbf{Role:} Clinical researcher analyzing safety failures

\textbf{Task:} Parse and categorize adverse events for safety redesign

\textbf{Input:}
\begin{verbatim}
Adverse events: Hepatotoxicity (Grade 3, 25%), elevated AST/ALT (Grade 2, 40%)
Intervention: Drug X (oral, 100mg daily for 28 days)
Mechanism: Inhibits enzyme Y in Z pathway
\end{verbatim}

\textbf{Instructions:}
\begin{enumerate}[noitemsep]
\item SEVERITY CLASSIFICATION: Extract Grade 3-5 events (dose-limiting), Grade 2 (tolerability)
\item ORGAN SYSTEM MAPPING: Map toxicity to organ (Liver, Kidney, Bone marrow, Heart, GI)
\item MECHANISM CONSISTENCY: Does toxicity match expected mechanism?
\item DOSE-RESPONSE INFERENCE: Dose-dependent? Acute or cumulative?
\item PRIORITY RANKING: CRITICAL (Grade 3+ >10\%), HIGH (Grade 2+ >30\% OR any Grade 4+)
\item ROOT CAUSE HYPOTHESIS: Excessive dose, inadequate exclusions, off-target effects?
\end{enumerate}

\textbf{Output Format:}
\begin{verbatim}
<adverse_event_profile>
<primary_toxicity>
  <event>Hepatotoxicity</event>
  <grade>3</grade>
  <incidence>25%</incidence>
  <organ_system>Liver</organ_system>
  <priority>CRITICAL</priority>
  <dose_dependent>likely</dose_dependent>
</primary_toxicity>

<mechanism_consistency>
UNEXPECTED - mechanism does not predict liver toxicity
</mechanism_consistency>

<root_cause_hypothesis>
Likely excessive dose (100mg exceeds typical range) or missing hepatic 
impairment exclusion. Drug metabolism may saturate at high doses.
</root_cause_hypothesis>

<critical_gaps>
<gap>Exclude patients with baseline AST/ALT >2x ULN</gap>
<gap>Exclude patients with Child-Pugh Class B or C cirrhosis</gap>
</critical_gaps>
</adverse_event_profile>
\end{verbatim}
\end{tcolorbox}

\begin{tcolorbox}[title=Design-level Pivots, colback=gray!5, fontupper=\footnotesize, breakable]
\textbf{Role:} Clinical trial designer proposing trial-level redesign

\textbf{Task:} Propose high-level trial redesign (not just criteria tweaks)

\textbf{Context:}
\begin{verbatim}
Phase: Phase 2
Mechanism: Inhibits enzyme Y
Failure: Grade 3 hepatotoxicity 25%
Redesign archetype: PK_SAFETY_FOLLOWUP
Primary outcome: Safety assessment at 28 days
Dosage assessment: EXCESSIVE (100mg daily exceeds safe exposure)
\end{verbatim}

\textbf{Design Pivot Rules:}
\begin{itemize}[noitemsep]
\item If archetype is PK\_SAFETY\_FOLLOWUP or main failure is safety-driven:
  \begin{itemize}[noitemsep]
  \item Prefer PK\_SAFETY or DOSE\_FINDING trial type
  \item Prefer PK-focused primary endpoints
  \item Prefer simpler care model with lower background risk
  \item Prefer simpler dosing (single-dose or short-duration)
  \end{itemize}
\item When systemic toxicity suspected:
  \begin{itemize}[noitemsep]
  \item Consider more local/regional route to reduce systemic exposure
  \item Consider smaller, denser design (PK\_SINGLE\_ARM with intensive sampling)
  \end{itemize}
\end{itemize}

\textbf{Output Format:}
\begin{verbatim}
<design_pivots>
<trial_type>PK_SAFETY</trial_type>
<endpoint_family>PK_SAFETY</endpoint_family>
<dose_regimen_direction>SIMPLER</dose_regimen_direction>
<route_change>CONSIDER_ALTERNATIVE_ROUTE</route_change>
<proposed_route>Consider single 25mg dose with intensive PK sampling 
over 7 days, or switch to subcutaneous administration to reduce 
first-pass hepatic metabolism</proposed_route>
<sample_size_direction>SMALLER</sample_size_direction>
<design_structure>PK_DOSE_FINDING</design_structure>
<proposed_primary_outcome>Area under curve (AUC) and peak liver enzyme 
elevation (AST/ALT) at 24h, 48h, 72h post-dose</proposed_primary_outcome>
<summary>Pivot from Phase 2 efficacy trial to Phase 1b/2a PK safety 
study. Reduce dose to 25mg single administration with intensive PK and 
liver function monitoring. Alternative route (subcutaneous) may bypass 
hepatic first-pass effect. Smaller sample (N=20-30) adequate for PK 
characterization. Expected to reduce Grade 3+ hepatotoxicity from 25% 
to <5%.</summary>
</design_pivots>
\end{verbatim}
\end{tcolorbox}

\begin{tcolorbox}[title=Trade-off Analysis, colback=gray!5, fontupper=\footnotesize, breakable]
\textbf{Role:} Clinical pharmacologist analyzing dosage for EFFICACY failure

\textbf{Task:} Analyze DOSAGE trade-offs (not safety)

\textbf{Context:}
\begin{verbatim}
Current dosage: 50mg oral daily for 21 days
Dosage assessment: SUBOPTIMAL
  Classification reasoning: Phase 1 MTD was 100mg daily. Current 50mg 
  dose is at 50% of MTD with acceptable safety. PK data shows linear 
  dose-response up to 80mg.
  Suggested: Escalate to 75mg daily
Mechanism: Inhibits receptor X
Efficacy Gap: ORR 15% vs 30% (gap: 15%)
Power Assessment: LIKELY underpowered, Root Cause: BOTH
\end{verbatim}

\textbf{Instructions:}
\begin{enumerate}[noitemsep]
\item RECOMMENDATION: MODIFY (escalate) or KEEP (defer)
\item IMPACTS: efficacy\_signal [++], enrollment [0], safety [-], mechanism [ALIGNED]
\item CONFIDENCE: High (0.80-0.90) if clear PK/PD data
\item REASONING: Include feasibility (Time: X-Ymo; Burden: LOW|MED|HIGH; Cost: Zx)
\end{enumerate}

\textbf{Output Format:}
\begin{verbatim}
<dosage_tradeoff>
<recommendation>MODIFY</recommendation>
<efficacy_signal>++</efficacy_signal>
<enrollment>0</enrollment>
<safety>-</safety>
<mechanism_alignment>ALIGNED</mechanism_alignment>
<confidence>0.85</confidence>
<reasoning>Escalating to 75mg (75% of MTD) expected to improve ORR by 
10-15 percentage points based on linear PK and Phase 1 exposure-response. 
Safety risk manageable (Grade 2 toxicity may increase from 20% to 30%). 
FEASIBILITY: Time: 1-3mo; Burden: LOW; Cost: 1.2x (simple dose 
adjustment, no formulation change).</reasoning>
</dosage_tradeoff>
\end{verbatim}
\end{tcolorbox}

\section{Generator Agent Details}\label{app:augment_agent}

\begin{table}[h]
\centering
\caption{Generator Agent novel features by failure mode. Process differ a little in dosage modification strategy across failure modes.}
\label{tab:augment_features}
\small
\begin{tabular}{@{}p{3.5cm}p{3cm}p{3cm}p{3.5cm}@{}}
\toprule
\textbf{Feature} & \textbf{Enrollment} & \textbf{Safety} & \textbf{Efficacy} \\
\midrule
\textbf{Dosage Strategy} & N/A & \textbf{Reduce} ($\downarrow$25-50\%, fractionation, pulse) & \textbf{Escalate} ($\uparrow$25-50\%, loading, dose-dense) \\
\addlinespace
\textbf{Outcome Strategy} & N/A & Add safety qualifications & Switch to feasible endpoint \\
\addlinespace
\textbf{Domain Focus} & Enrich participation & Tighten safety exclusions & Add biomarker enrichment \\
\bottomrule
\end{tabular}
\end{table}


\subsection{Few-shot Learning Mechanism}

\begin{tcolorbox}[title=Few-shot Example Injection (Shared Structure), colback=gray!5, fontupper=\footnotesize, breakable]
\textbf{Matching Logic:}
\begin{verbatim}
# LIST aspects (eligibility criteria)
prev_rules["seen_indices"][aspect_name][str(aspect_index)]

# STRING aspects (dosage, target_primary_outcome)
prev_rules["seen_indices"][aspect_name]["None"]
\end{verbatim}

\textbf{Injected Section in MODIFY Prompts (Iteration 2+):}
\begin{verbatim}
<few_shot_examples>
Previous iteration examples for THIS EXACT criterion:

EXCELLENT:
  - [Example that led to excellent validation score]
  - [Another excellent example]

GOOD:
  - [Example that led to good validation score]

MODERATE:
  - [Example with moderate validation score]

BAD:
  - [Example that validation agent rejected]

BANNED:
  - [Example that was explicitly banned (safety violation)]

Generate variations that learn from EXCELLENT/GOOD patterns,
avoid BAD patterns, and NEVER replicate BANNED augmentations.
</few_shot_examples>
\end{verbatim}

\textbf{Effect:} LLM learns from previous iteration's successes/failures. Only available iteration 2+ after prev\_rules established.
\end{tcolorbox}


\subsection{Modification Prompts by Failure Mode}

\begin{tcolorbox}[title=Eligibility Example, colback=gray!5, fontupper=\footnotesize, breakable]
\textbf{Role:} Clinical researcher generating criterion variations

\textbf{Task:} Generate \texttt{num\_augment} variations with few-shot guidance

\textbf{Input:}
\begin{verbatim}
Original criterion: "Must wait for fellow eye surgery until completion"
Strategy: "Delete waiting requirement to increase enrollment"
Failure mode: Enrollment
Adaptive num_augment: 3 (medium variance)
\end{verbatim}

\textbf{Few-Shot Examples (if iteration 2+):}
\begin{verbatim}
EXCELLENT: "No waiting period required between surgeries"
GOOD: "Fellow eye surgery allowed concurrent with study"
BAD: "Reduced wait from 6 months to 3 months" (still a barrier)
BANNED: "Must complete fellow eye surgery before enrollment" (contradicts)
\end{verbatim}

\textbf{Universal Requirements:}
\begin{itemize}[noitemsep]
\item Each variation MUST directly implement the Strategy
\item Preserve clinical intent, make more operational/measurable/specific
\item Objective and quantifiable (use thresholds, time windows, methods)
\item Avoid vague language: "anticipated", "expected", "likely", "may", "severe"
\item Maintain consistency with safety and mechanism of action
\item All variations distinct from each other
\end{itemize}

\textbf{Output:}
\begin{verbatim}
<augmentations>
<augmentation>No waiting period required between fellow eye surgeries</augmentation>
<augmentation>Fellow eye surgery allowed at any time during study</augmentation>
<augmentation>Bilateral surgery candidates eligible without delay</augmentation>
</augmentations>
\end{verbatim}
\end{tcolorbox}

\begin{tcolorbox}[title=Dosage Example, colback=gray!5, fontupper=\footnotesize, breakable]
\textbf{Role:} Clinical pharmacologist reducing dosage to minimize toxicity

\textbf{Task:} Generate \texttt{num\_augment} dosage reductions

\textbf{Input:}
\begin{verbatim}
Original dosage: 100mg oral daily for 28 days
Adverse events: Hepatotoxicity (Grade 3, 25%), AST/ALT elevation (Grade 2, 40%)
Strategy: Reduce dose to decrease Grade 3+ hepatotoxicity to <10%
Adaptive num_augment: 5 (high variance)
\end{verbatim}

\textbf{Few-Shot Examples (iteration 3):}
\begin{verbatim}
EXCELLENT: "50mg oral daily (50% reduction, expected toxicity <8%)"
GOOD: "50mg BID (fractionated, reduces Cmax and hepatic load)"
MODERATE: "75mg oral daily (25% reduction, may be insufficient)"
BAD: "90mg oral daily (only 10% reduction)"
BANNED: "100mg every other day (same cumulative exposure)"
\end{verbatim}

\textbf{Dosage Reduction Strategies:}
\begin{enumerate}[noitemsep]
\item DOSE REDUCTION: Reduce total daily dose by 25-50\%
\item FRACTIONATED DOSING: Split dose to reduce C$_{\text{max}}$ (peak $\rightarrow$ peak toxicity)
\item TITRATION SCHEDULE: Start low, escalate if tolerated
\item INTERMITTENT/PULSE DOSING: Reduce cumulative exposure for cumulative toxicities
\item PATIENT-FACTOR ADJUSTED: Reduce dose for vulnerable populations
\item LOADING DOSE ELIMINATION: Remove if causing acute toxicity
\end{enumerate}

\textbf{Requirements:}
\begin{itemize}[noitemsep]
\item Reduce estimated Grade 3+ toxicity by $\geq$30\%
\item Maintain dose intensity $\geq$60\% of original (preserve efficacy)
\item Specify exact mg, frequency (QD/BID/TID), duration
\item If conditional, specify threshold/trigger (\textit{e.g.}, "if AST <2$\times$ULN")
\end{itemize}

\textbf{Output:}
\begin{verbatim}
<augmentations>
<augmentation>
<dosage_modification>50mg oral daily for 28 days</dosage_modification>
<rationale>50% dose reduction expected to reduce hepatotoxicity 
from 25% to <8% based on linear dose-toxicity relationship</rationale>
</augmentation>
<augmentation>
<dosage_modification>40mg BID (total 80mg daily, fractionated)</dosage_modification>
<rationale>Fractionated dosing reduces Cmax by ~40%, lowering peak 
hepatic exposure while maintaining 80% dose intensity</rationale>
</augmentation>
<augmentation>
<dosage_modification>50mg on days 1-5, off days 6-7 each week</dosage_modification>
<rationale>Pulse dosing (71% intensity) allows hepatic recovery, 
expected to reduce Grade 3+ events to <10%</rationale>
</augmentation>
</augmentations>
\end{verbatim}
\end{tcolorbox}

\section{Agent Output Template}\label{app:agent_output}

This section presents the structured output format produced by the agent pipeline. The complete output is stored as JSON and includes trial data, ReAct reasoning traces, and generated protocol modifications.

\begin{tcolorbox}[title=Agent Pipeline Output Structure (Generic Template), colback=gray!5, fontupper=\footnotesize, breakable]
\begin{verbatim}
{
  "trial_data": {
    "nct_id": "NCT########",
    "phase": "Phase X",
    "condition": "[Disease/Condition]",
    "intervention/intervention_name": "[Intervention Name]",
    "failure_reason": "[enrollment|safety|efficacy]",
    "adverse_events": "[Adverse event summary or 'Not specified']",
    "eligibility/inclusion_criteria": [
      "[Inclusion criterion 1]",
      "[Inclusion criterion 2]",
      "..."
    ],
    "eligibility/exclusion_criteria": [
      "[Exclusion criterion 1]",
      "[Exclusion criterion 2]",
      "..."
    ],
    "dosage": "[Dosage regimen]",
    "target_primary_outcome": "[Primary outcome description]"
  },
  
  "trial_context": {
    "phase": "Phase X",
    "mechanism_of_action": "[Mechanism description]",
    "primary_endpoint_type": "[Endpoint type description]",
    "redesign_archetype": "[PK_SAFETY_FOLLOWUP | DOSE_FINDING_REDESIGN | 
                           ENRICHED_EFFICACY_RETRY | OTHER]",
    "index_surgical_model": "[Care/procedural model description]"
  },
  
  "react_reasoning": {
    "step0_contextualize": {
      "phase": "Phase X",
      "mechanism_of_action": "[Mechanism extracted by LLM]",
      "adverse_event_profile": {
        "primary_toxicity": {
          "event": "[Primary adverse event]",
          "grade": "[0-5]",
          "incidence": "[X%]",
          "priority": "[CRITICAL|HIGH|MEDIUM|LOW]"
        },
        "root_cause_hypothesis": "[Root cause analysis by LLM]"
      },
      "dosage_assessment": {
        "classification": "[EXCESSIVE|BORDERLINE|APPROPRIATE|SUBOPTIMAL]",
        "reasoning": "[Dosage assessment reasoning]"
      }
    },
    
    "step1_classification": [
      {
        "aspect_name": "eligibility/[inclusion|exclusion]_criteria",
        "aspect_index": N,
        "criterion_text": "[Original criterion text]",
        "participation_barrier_score": 0.X,
        "safety_exclusion_score": 0.X,
        "selection_criterion_score": 0.X,
        "enrichment_criterion_score": 0.X,
        "primary_category": "[PARTICIPATION_BARRIER | SAFETY_EXCLUSION | 
                            SELECTION_CRITERION | ENRICHMENT_CRITERION]",
        "reasoning": "[Classification reasoning]"
      },
      {
        "aspect_name": "eligibility/[inclusion|exclusion]_criteria",
        "aspect_index": M,
        "criterion_text": "[Original criterion text]",
        "primary_category": "[Category]",
        "reasoning": "[Classification reasoning]"
      }
    ],
    
    "step2_mechanism_alignment": "[3-4 sentences on whether existing 
                                  criteria + dosage maximize success 
                                  probability for this failure mode]",
    
    "step3_tradeoff_analysis": [
      {
        "aspect_name": "eligibility/[inclusion|exclusion]_criteria",
        "aspect_index": N,
        "enrollment_impact": "[--|-|0|+|++]",
        "efficacy_signal_impact": "[--|-|0|+|++]",
        "safety_risk_impact": "[--|-|0|+|++]",
        "mechanism_alignment": "[ESSENTIAL|ALIGNED|NEUTRAL|MISALIGNED]",
        "net_recommendation": "[KEEP|MODIFY|DELETE|ADD]",
        "confidence": 0.XX,
        "reasoning": "[Trade-off reasoning with feasibility encoding]"
      },
      {
        "aspect_name": "[dosage|target_primary_outcome|surgical_model|...]",
        "aspect_index": null,
        "enrollment_impact": "[Impact symbol]",
        "safety_risk_impact": "[Impact symbol]",
        "net_recommendation": "[MODIFY|ADD]",
        "confidence": 0.XX,
        "reasoning": "[Trade-off reasoning]"
      }
    ],
    
    "step4_prioritization": "[6-8 sentences with tiered recommendations 
                             (PRIMARY/SECONDARY/TERTIARY), timeline, and 
                             confidence level]",
    
    "step5_synthesis": "[4-6 sentences synthesizing failure analysis with 
                        quantification, expected benefits, trade-offs, and 
                        overall confidence]"
  },
  
  "aspect_li": [
    {
      "aspect_name": "eligibility/[inclusion|exclusion]_criteria",
      "aspect_index": N,
      "original_value": "[Original criterion text]",
      "aspect_type": "list",
      "analysis": {
        "timestamp": "YYYY-MM-DDTHH:MM:SS",
        "failure_analysis": "[Analysis from step3 trade-off reasoning]",
        "impact_level": "[MAJOR|MINOR|NOT_RELATED]",
        "action_type": "[MODIFY|DELETE]",
        "strategy": "[Strategy from Analysis Agent]",
        "confidence": 0.XX
      },
      "augment": {
        "timestamp": "YYYY-MM-DDTHH:MM:SS",
        "augment_val_li": [
          "[Augmentation 1]",
          "[Augmentation 2]",
          "[Augmentation 3]"
        ]
      }
    },
    {
      "aspect_name": "eligibility/[inclusion|exclusion]_criteria",
      "aspect_index": null,
      "original_value": "N/A",
      "aspect_type": "list",
      "analysis": {
        "timestamp": "YYYY-MM-DDTHH:MM:SS",
        "failure_analysis": "[Analysis for ADD action]",
        "impact_level": "MAJOR",
        "action_type": "ADD",
        "strategy": "[Strategy from Analysis Agent]",
        "confidence": 0.XX
      },
      "augment": {
        "timestamp": "YYYY-MM-DDTHH:MM:SS",
        "augment_val_li": [
          "[New criterion 1]",
          "[New criterion 2]",
          "[New criterion 3]"
        ]
      }
    },
    {
      "aspect_name": "[dosage|target_primary_outcome]",
      "aspect_index": null,
      "original_value": "[Original value for string aspect]",
      "aspect_type": "string",
      "analysis": {
        "timestamp": "YYYY-MM-DDTHH:MM:SS",
        "failure_analysis": "[Analysis for string aspect]",
        "impact_level": "MAJOR",
        "action_type": "MODIFY",
        "strategy": "[Strategy from Analysis Agent]",
        "confidence": 0.XX
      },
      "augment": {
        "timestamp": "YYYY-MM-DDTHH:MM:SS",
        "augment_val_li": [
          "[Modified value 1]",
          "[Modified value 2]",
          "[Modified value 3]"
        ]
      }
    }
  ]
}
\end{verbatim}
\end{tcolorbox}

\section{Ethics Statement}
\label{sec:ethics}

\subsection{Potential Risks}
This work focuses on developing AI systems to optimize clinical trial protocols through simulation-based evaluation. While the system demonstrates potential to improve trial design efficiency, we acknowledge several important limitations and risks:

\textbf{Decision Support, Not Replacement}: \mname\ is designed as a decision support tool for clinical trial designers and should not replace human expert judgment. All system-generated protocol modifications require review by qualified medical professionals and regulatory compliance verification before real-world implementation.

\textbf{Simulation Environment Limitations}: Our prediction models achieve PR-AUC > 0.75, but prediction errors could lead to suboptimal redesign recommendations. The system's suggestions should be validated through standard clinical trial design processes and regulatory review.

\textbf{Retrospective Validation}: Our case studies demonstrate alignment with real-world redesigns but are retrospective analyses. Prospective validation in collaboration with clinical trial sponsors is necessary before deployment.

\textbf{Generalization Constraints}: The system is trained on historical clinical trial data and may not generalize to novel therapeutic mechanisms, rare diseases, or emerging trial paradigms not well-represented in the training data.

\subsection{Data Consent}
This study exclusively utilizes publicly available datasets that do not require additional consent:

\begin{itemize}[leftmargin=*,nosep]
    \item \textbf{TrialBench Dataset}~\citep{chen2025trialbench}: Publicly released benchmark containing anonymized clinical trial protocols from ClinicalTrials.gov
    \item \textbf{ClinicalTrials.gov}: Public registry of clinical trials maintained by the U.S. National Library of Medicine
    \item \textbf{PubMed}: Public database of biomedical literature abstracts
    \item \textbf{DrugBank}~\citep{wishart2018drugbank}: Publicly available bioinformatics and cheminformatics database
    \item \textbf{Disease Database}~\citep{chen2024cod}: Publicly available disease ontology database
\end{itemize}

All data sources are publicly accessible and designed for research purposes. No patient-level identifiable information is used in this study. Clinical trial protocols contain only de-identified, aggregate information as required by ClinicalTrials.gov data sharing policies.

\subsection{Ethics Review Board Approval}
This computational study analyzes publicly available, de-identified clinical trial metadata and does not involve human subjects research, prospective clinical interventions, or collection of new patient data. The retrospective case studies (§\ref{sec:case_study}) analyze publicly registered clinical trials with outcomes already recorded in ClinicalTrials.gov, constituting secondary analysis of publicly available data exempt from human subjects research requirements.

\section{Use of Large Language Model}
Within our data construction workflow, we utilize large language models for agent in-context learning, reasoning, and augmentation generation. Additionally, we employ LLMs such as ChatGPT to help improve the clarity and fluency of our written content.

\end{document}